# Lacking Data? No worries! How synthetic images can alleviate image scarcity in wildlife surveys: a case study with muskox (*Ovibos moschatus*)

## Simon Durand[1,4], Samuel Foucher[1], Alexandre Delplanque[2], Joëlle Taillon[3], Jérôme Théau*[1,4]

[1]Department of Applied Geomatics, Université de Sherbrooke, 2500 Boulevard de l'Université, Sherbrooke QC, J1K 2R1, Canada (*corresponding author: jerome.theau@usherbrooke.ca)

[2]TERRA Teaching and Research Centre (Forest Is Life), Uliège, Gembloux Agro-Bio Tech, 2 Passage des Déportés, Gembloux 5030, Belgium

[3]Direction générale de la gestion de la faune, Ministère de l'Environnement, de la Lutte contre les Changements Climatiques, de la Faune et des Parcs (MELCCFP), 880 Chemin Sainte-Foy, Québec QC, G1S 4X4, Canada

[4]Quebec Centre for Biodiversity Science (QCBS), Stewart Biology, McGill University, Montréal QC, H3A 1B1, Canada

## Abstract

Accurate population estimates are essential for wildlife management, providing critical insights into species abundance and distribution. Traditional survey methods, including visual aerial counts and GNSS telemetry tracking, are widely used to monitor muskox (*Ovibos moschatus*) populations in Arctic regions. These approaches are resource-intensive and constrained by logistical challenges. Advances in remote sensing, artificial intelligence, and high-resolution aerial imagery offer promising alternatives for wildlife detection. Yet, the effectiveness of deep learning object detection models (ODMs) is often limited by small datasets, making it challenging to train robust ODMs for sparsely distributed species like muskoxen. This study investigates the integration of synthetic imagery, created with diffusion-based models, to supplement limited training data and improve muskox detection in zero-shot and few-shot settings.

We compared a baseline model trained solely on real imagery with five zero-shot (ZS1-ZS5) and five few-shot (FS1-FS5) models that incorporated progressively more synthetic imagery in the training set. For the zero-shot models, where no real images were included in the training set, adding synthetic imagery improved detection performance. As more synthetic images were added, performance in precision, recall and F1 score increased, but eventually plateaued, suggesting diminishing returns when synthetic images exceeded 100% of the baseline model training dataset. For few-shot models, combining real and synthetic images led to better recall and slightly higher overall accuracy compared to using real images alone, though these improvements were not statistically significant.

Our findings demonstrate the potential of synthetic images to train accurate ODMs when data is scarce, offering important perspectives for wildlife monitoring by enabling rare or inaccessible species to be monitored and to increase monitoring frequency. This approach could be used to initiate ODMs without real data and refine it as real images are acquired over time.

## 1. Introduction

The development of wildlife management and conservation plans typically relies on accurate estimates of animal density or abundance and population trends based on direct or indirect survey methods (Lancia et al., 2005; McComb et al., 2010; Sinclair et al., 2006). By providing critical data on animal populations, wildlife surveys play a crucial role in assessing the health of habitats, identifying threatened species, understanding population dynamics, monitoring the influence of human activities on biodiversity, and studying changes in the spatial distribution of species, which are essential for developing effective management strategies (Buckland et al., 2001; McComb et al., 2010; Walters, 1986; Witmer, 2005).

Traditional wildlife monitoring methods, such as total and sampled counts as well as indirect estimates of population size (e.g., footprints, droppings, nests), are resource-intensive which may limits survey frequency and geographical scope (Newey et al., 2015; Prosekov et al., 2020; Sinclair et al., 2006; Wang et al., 2019; Witmer, 2005). In addition, indirect methods using indices may not accurately estimate population size due to imperfect metric correlations (Lancia et al., 2005; Prosekov et al., 2020; Williams et al., 2002).

Distance sampling conducted through aerial surveys, typically using visual counting techniques, is a common approach among wildlife management organizations for estimating large terrestrial animal abundance (Le Moullec et al., 2017; Prosekov et al., 2020; Schmidt et al., 2022). These methods are highly efficient in covering large and open areas (Lancia et al., 2005) but are not without limitations, including observer fatigue, reduced visibility in adverse weather, and the need for trained personnel (Caughley, 1974; Fleming and Tracey, 2008; Witmer, 2005). In recent years, distance sampling has evolved to incorporate the photographing of large populations, which enhances estimates by allowing for more accurate post-survey counting and detailed population composition analyses, such as assessments of age and sex ratios (Adamczewski et al., 2021; Couturier et al., 2018; Davison and Williams, 2022; Schlossberg et al., 2016). However, the application of these improved hybrid methods to species like the muskox (*Ovibos moschatus*), which are unevenly and sparsely distributed across vast and remote territories, presents additional challenges and complexities.

The muskox is a large ungulate belonging to the Bovidae family, primarily inhabiting the Canadian Arctic tundra and Greenland (Cuyler et al., 2020; Gunn and Adamczewski, 2003). This species holds significant economic and cultural value for Inuit communities, serving as both a traditional food source and a contributor to local economies through hunting and qiviut fleece (Cuyler et al., 2020; Kutz et al., 2017). Most muskox populations are welcomed by indigenous communities for their integral role in socio-cultural identity, artistic expressions, and traditional diet. In regions where muskox presence is recent or where the species has been introduced, however, muskox are viewed as a potential competitor for declining caribou (*Rangifer tarandus*) populations (Brodeur et al., 2023; Kutz et al., 2017). Moreover, climate change is increasingly affecting the Arctic, leading to more frequent and intense winter warming and rain-on-snow events (AMAP, 2017). These changes significantly disrupt the foraging habits of species such as caribou and muskoxen, while also altering vegetation through processes like shrubification which can have a variable impact on herbivore species (AMAP, 2017; Lemay et al., 2018). This highlights the need for accurate population monitoring to assess the health and resilience of wildlife populations and the ecosystems they inhabit, as well as to promote sustainable coexistence between species such as muskoxen and local communities (Cuyler et al., 2020; Kutz et al., 2017).

Distance sampling, most commonly conducted through aerial transect surveys, is the main method used to estimate muskoxen populations across their Arctic range (Cuyler et al., 2020; Gunn and Adamczewski, 2003). These surveys are typically carried out between March and September, with a team of observers counting animals and recording compositional characteristics using either helicopters or small planes (Cuyler et al., 2020). In this context, oblique imagery is typically preferred during visual surveys, as it facilitates identification of animal and assessment of sex and age (Adamczewski et al., 2021; Lamprey et al., 2020).

Estimating abundance is a complex process due to numerous challenges in data collection. Conducting surveys in the Arctic is costly and resource-intensive (Kutz et al., 2017), requiring trained personnel such a pilots, observers, and data analysts, along with specialized equipment like cameras and Global Navigation Satellite System (GNSS) beacons (Alaska Department of Fish and Game, 2001; Anderson, 2016; Davison and Williams, 2022). The logistical challenges are compounded by the remoteness of wildlife habitats, necessitating refueling stations, favorable weather conditions, and adherence to regulations, all while trying to cover vast areas sometimes crossing jurisdictional boundaries with low and uneven animal density (Cuyler et al., 2020; Environment and Climate Change Canada, 2022).

These challenges contribute to the complexity of monitoring muskox populations and partly explain the absence or imprecision of key demographic parameters for several populations (Cuyler et al., 2020; Kutz et al., 2017). This lack of information limits evidence-based management decisions and constrains the broader understanding of muskox ecology, as well as their economic potential for northern communities (Kutz et al., 2017).

Recent advances in drone technology and sensor miniaturization, combined with the increasing availability of high-resolution satellite imagery have the potential of revolutionizing wildlife detection (Delplanque et al., 2024b; Linchant et al., 2015; Wang et al., 2019). These innovations, when combined with artificial intelligence, particularly convolutional neural networks (CNNs), offer numerous benefits such as decreased costs and time invested, reduced risk related with working in remote regions, and automated data processing (Delplanque et al., 2023a; Eikelboom et al., 2019; Gonzalez et al., 2016). Despite these benefits, developing such approaches requires large datasets, which can be a limitation for species like the muskox, where data is scarce.

To address this limitation, zero-shot and few-shot learning techniques have been developed, enabling models to rapidly generalize to new tasks from limited data (Antonelli et al., 2022; Huang et al., 2023; Wang et al., 2021). These approaches empower models to learn and adapt to new tasks or domains with minimal annotated samples, facilitating more precise and efficient monitoring (Yang et al., 2022). Key strategies within zero-shot and few-shot learning include metric-based, optimization-based, memory-based, data augmentation-based, and transfer learning methods (Wang et al., 2021). While few-shot strategies have been applied in certain ecological studies (Yang et al., 2022; Q. Zhang et al., 2023), the use of synthetic data, particularly synthetic images created with advanced generative models, as part of data augmentation or zero-shot frameworks remains unexplored in the context of wildlife surveys.

The objective of this study is to assess the effectiveness of zero-shot and few-shot learning approaches in automating the detection and counting of remote and sparsely distributed species such as muskoxen using both real and synthetic high-resolution aerial imagery. The synthetic images were produced using advanced generative methods, such as the diffusion model DALL-E 2, to simulate realistic environmental conditions and animal representations.

## 2.   Materials and Methods

### 2.1.   Study Area

Study areas are located in Quebec and the Northwest Territories (Canada), where nadir-like aerial imagery of muskoxen has been acquired (Figure 1). All images are considered nadir-like, as they were captured with a look angle of $0° \pm 30°$. In northern Quebec, wild ranging populations of muskoxen live in the Ungava Bay and Hudson Bay regions (Brodeur et al., 2023). These regions are dominated by barren-cuestas, wetlands, erect-shrub tundra, prostrate-shrub tundra and shrublands (Leboeuf et al., 2018). The Zoo Sauvage de Saint-Félicien, located in the Saguenay-Lac-Saint-Jean region of Quebec, houses seven muskoxen in captivity in a fenced natural environment with variable vegetation cover ranging from boreal forest to open environments. In the Northwest Territories, wild populations of muskoxen are found around the East Arm of Great Slave Lake, an area south of the treeline in the northern boreal forest. This region is characterized by its many islands, clear waters, and deep, complex shorelines (Piper, 2016).

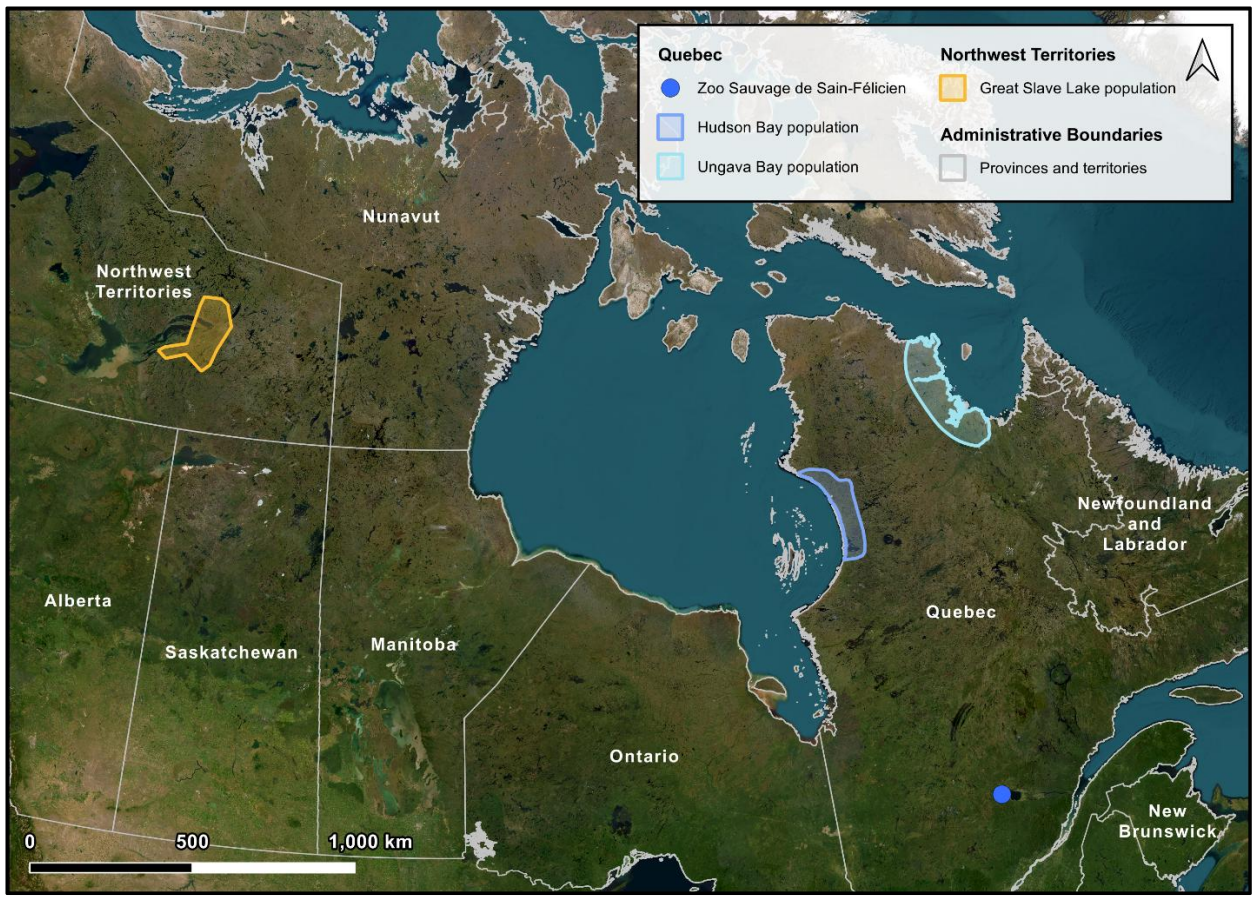

**Figure 1.** Map of muskoxen population areas with acquired aerial imagery in Quebec and the Northwest Territories (Canada).

### 2.2.   Methodology

The methodological approach (Figure 2) consists of two main phases. The first phase, data preparation, includes the acquisition of nadir-like aerial images of muskoxen and the creation of synthetic images to supplement the training datasets. Both real and synthetic images were annotated. The second phase focuses on training multiple CNN-based object detection models (ODMs) to automatically detect and count muskoxen in these nadir-like images. ODMs inspired by zero-shot learning strategies were trained exclusively on synthetic images, while ODMs inspired by few-shot learning strategies were trained on datasets that combined real and synthetic images. Additionally, classical data augmentation techniques were applied to all training datasets to further diversify them. Each of these steps is detailed in the following subsections of the methodology.

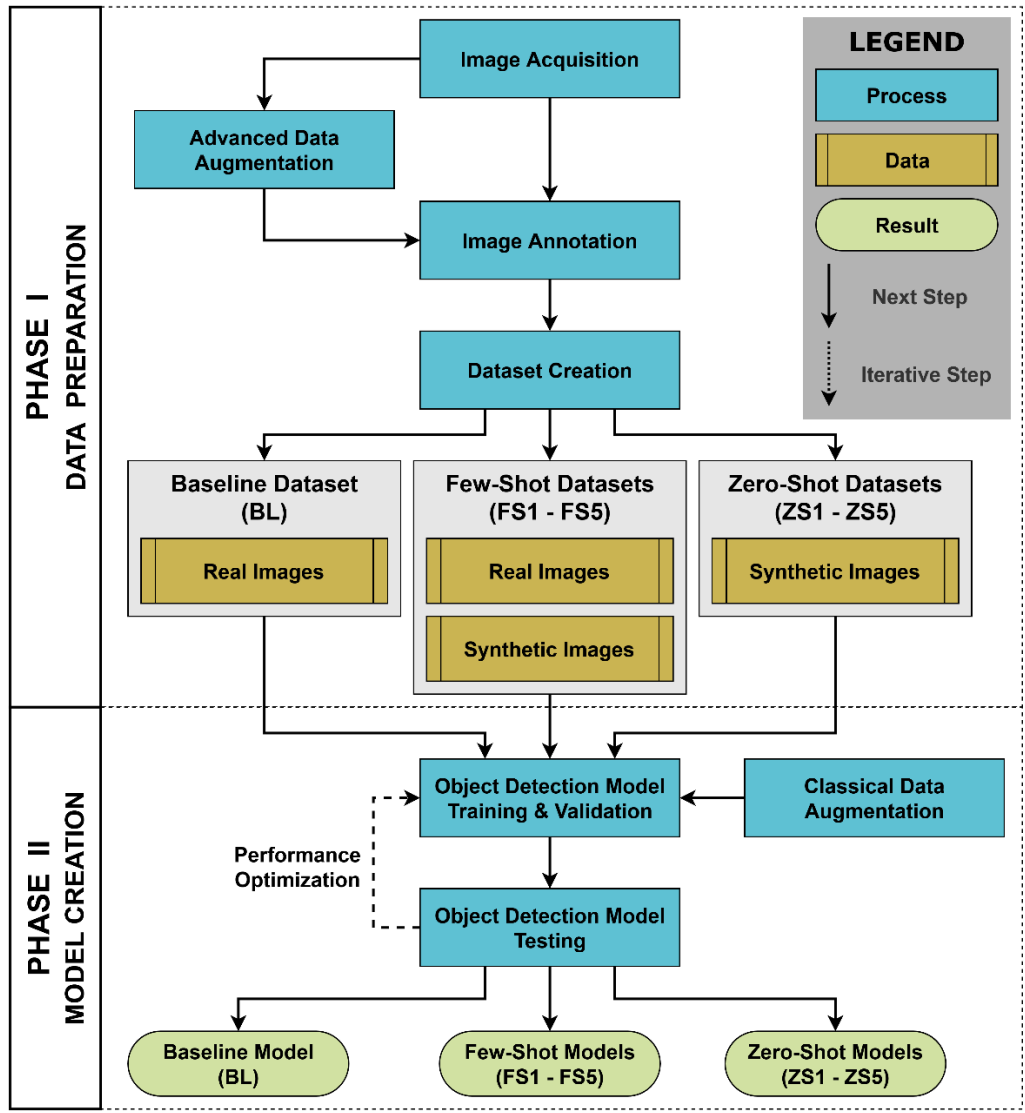

**Figure 2.** Methodological flowchart for the development of the object detection models.

### 2.2.1. Nadir-Like Aerial Imagery Acquisition

Aerial nadir-like images used in this study were provided by two government agencies and a wildlife zoo. The *Ministère de l'Environnement, de la Lutte contre les Changements Climatiques, de la Faune et des Parcs* (MELCCFP) of Quebec government and the Department of Environment and Natural Resources of the Northwest Territories provided images from previous wildlife surveys (Table 1). Additionally, a field campaign at the Zoo Sauvage de Saint-Félicien in 2022 yielded further aerial images captured by drone over the muskox enclosure. In total, 96 nadir-like aerial images were used for training and validation of the ODMs, while the 996 nadir-like aerial images acquired by the MELCCFP in 2023 were reserved for testing (Table 1). This data distribution was designed to emphasize ODM training based on scarce data, and testing using a dedicated field campaign aimed to acquire nadir-like images.

**Table 1.** Specifications of acquired nadir-like aerial images. All images are nadir-like, meaning they were acquired with a look angle of $0° \pm 30°$.

|  | Training & Validation | | | Testing |
| --- | --- | --- | --- | --- |
| Dataset | Zoo Sauvage de Saint-Félicien | Hudson Bay, Ungava Bay | Great Slave Lake | Hudson Bay, Ungava Bay |
| Study area | Quebec, Canada | Quebec, Canada | Northwest Territories, Canada | Quebec, Canada |
| Source | Cégep de Saint-Félicien | MELCCFP | ENR | MELCCFP |
| Acquisition period | March, 2022 | March, 2019-20 | March-April, 2018 | March, 2023 |
| Aerial platform | Drone | Helicopter | Plane | Helicopter |
| Flight altitude | 60-100 m | 100-400 m | 100-500 m | 100-400 m |
| Camera model | FC6310R | Canon EOS 5DS | Nikon D3X, Nikon D800 | Canon EOS 5DS |
| Zoom lens [1] | Phantom 4 Pro (DJI), 8,8-24 mm, f/2,8-11 | EF (Canon), 100-400 mm, f/4,5-5,6 | Nikkor AF-S (Nikon), 28-300 mm, f/3,5-5,6 | EF (Canon), 100-400 mm, f/4,5-5,6 |
| Image resolution | 4,864 x 3,648 (17.74 MP) | 8,688 x 5,792 (50.32 MP) | 7,360 x 4,912 (36.15 MP) | 8,688 x 5,792 (50.32 MP) |
| Ground Sampling Distance | 0.68-3.08 cm/pixel | 0.10-1.66 cm/pixel | 0.16-8.71 cm/pixel | 0.10-1.66 cm/pixel |
| Image quantity | 60 | 35 | 1 | 996 |

[1] The information presented includes the name and brand of the lens, focal length and aperture.

ENR: Department of Environment and Natural Resources

MELCCFP: *Ministère de l'Environnement, de la Lutte contre les Changements Climatiques, de la Faune et des Parcs*

### 2.2.2. Advanced Data Augmentation

Advanced data augmentation involved the creation of synthetic images to enhance the diversity of data used for training ODMs, leveraging OpenAI's DALL-E 2 text-to-image diffusion model (Ramesh et al., 2022). The model created synthetic images based on text prompts describing the muskox's ecological context (e.g., "snowy landscape with herds of up to 75 animals") and specific image acquisition characteristics (e.g., "nadir-like view, photorealistic style"). Among several prompts tested (see Appendix S1), "Herd of muskoxen seen from above with a winter background, aerial imagery" consistently produced the best results. Images were generated via a custom Python script using the DALL-E 2 API. This allowed for automated batch creation of 10 images at a resolution of 1024 x 1024 pixels to allow later cropping and resizing when incomplete muskoxen appeared on the image periphery. Synthetic images were produced between 2022 and 2024, after which they were manually filtered to retain only the most realistic outputs, resulting in a total of 160 retained images (Figure 3). This approach was used to augment datasets for both zero-shot and few-shot models.

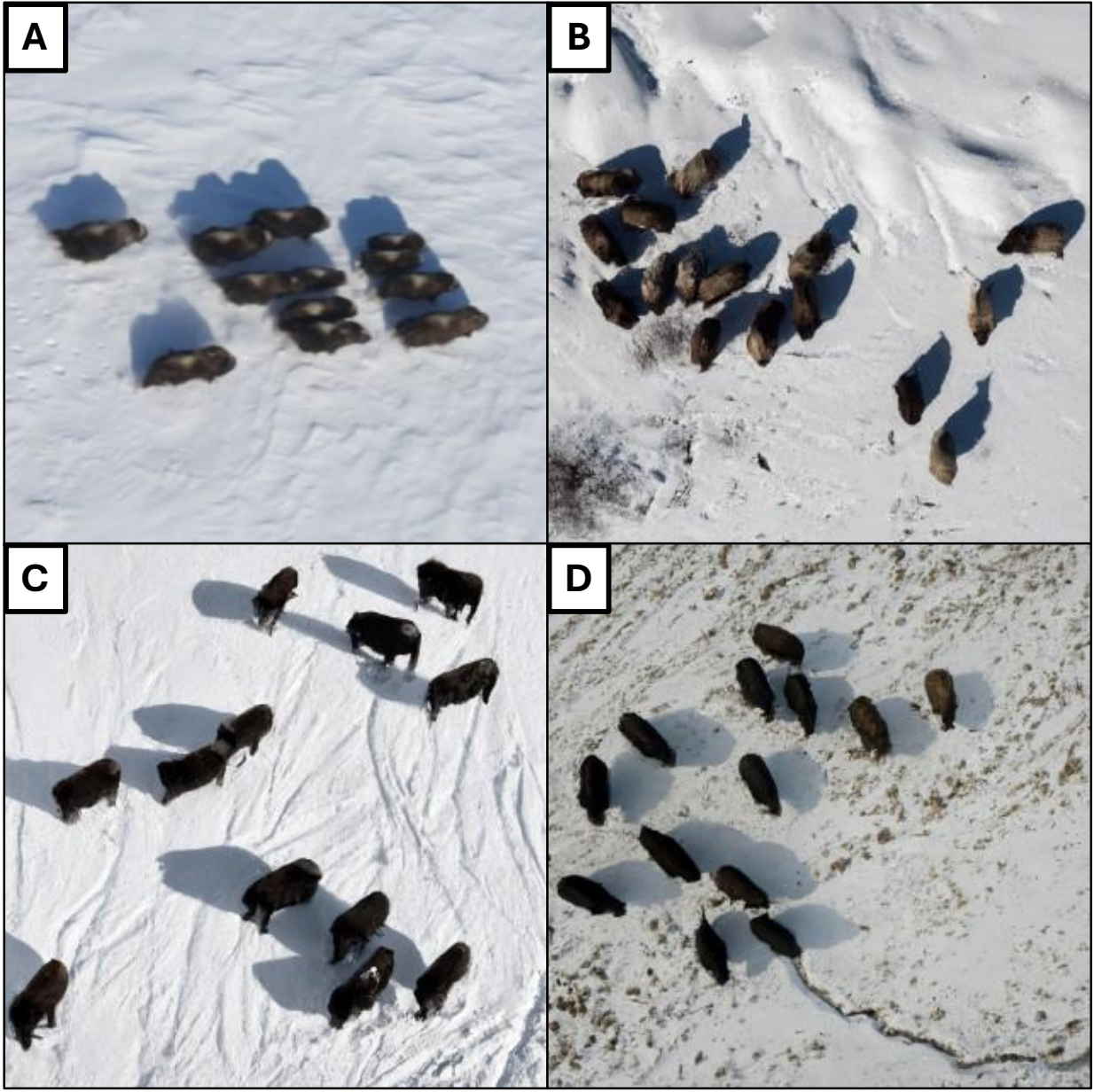

**Figure 3.** Examples of real aerial imagery [A] and synthetic aerial imagery [B, C and D] produced using DALL-E 2. The synthetic images were generated with the text prompt "Herd of muskoxen seen from above with a winter background, aerial imagery".

### 2.2.3. Dataset Creation

Eleven separate datasets were created to assess the effectiveness of zero-shot and few-shot learning techniques using synthetic images (Table 2). All except the baseline dataset included a combination of real and synthetic nadir-like aerial images. The baseline dataset, which consists only of real images, serves as a baseline to benchmark performance against the zero-shot (ZS1–ZS5) and few-shot (FS1–FS5) datasets. The zero-shot datasets were composed exclusively of synthetic images. For the few-shot datasets, all real images were included, with synthetic images progressively added to increase the ratio of synthetic-to-real data. Each dataset was partitioned into training (80%) and validation (20%) subsets, with testing using the dedicated subset of 996 images. All images were manually annotated by a member of the research team using the open-source data labeling tool Label Studio (Tkachenko et al., 2020).

**Table 2.** Composition of datasets. The baseline (BL) dataset exclusively comprises real images. The zero-shot datasets (ZS1-ZS5) consist solely of synthetic images. The few-shot datasets (FS1-FS5) encompass a combination of real and synthetic images.

| Images | Datasets | | | | | | | | | | |
|---|---|---|---|---|---|---|---|---|---|---|---|
| | BL | ZS1 | ZS2 | ZS3 | ZS4 | ZS5 | FS1 | FS2 | FS3 | FS4 | FS5 |
| Real | 96 | 0 | 0 | 0 | 0 | 0 | 96 | 96 | 96 | 96 | 96 |
| Synthetic | 0 | 30 | 60 | 96 | 130 | 160 | 30 | 60 | 96 | 130 | 160 |
| Total | 96 | 30 | 60 | 96 | 130 | 160 | 126 | 156 | 192 | 226 | 256 |

Subsequently, the images were then decomposed into patches with a resolution of 512 x 512 pixels, preserving only those featuring muskoxen (Table 3). To ensure uniformity in muskox size across the images, resizing was performed prior to patch creation. This step is essential to account for different flight altitudes and variations in equipment used during image capture.

**Table 3.** Number of patches and labels available for training and validation of object detection models across datasets. Due to their limited dimensions, synthetic images provided significantly fewer patches for zero-shot models (ZS1-ZS5) compared to those available for few-shot models (FS1-FS5) and the baseline (BL) model.

| | Datasets | | | | | | | | | | |
|---|---|---|---|---|---|---|---|---|---|---|---|
| | BL | ZS1 | ZS2 | ZS3 | ZS4 | ZS5 | FS1 | FS2 | FS3 | FS4 | FS5 |
| Patches | 1,189 | 183 | 338 | 544 | 741 | 910 | 1,372 | 1,527 | 1,733 | 1,930 | 2,099 |
| Animals | 4,083 | 669 | 1,148 | 1,829 | 2,571 | 3,042 | 4,752 | 5,231 | 5,912 | 6,654 | 7,125 |

### 2.2.4. Classical Data Augmentation

Classical data augmentation strategies involve leveraging image transformation operations to augment the dataset, primarily to limit overfitting when the training dataset is small and to improve the ODM's generalization. The Python library *Albumentations* (Buslaev et al., 2020) was used to implement these operations. Throughout the training and validation of the ODM, image normalization was applied alongside operations such as brightness and contrast adjustments, hue and saturation variations, flipping, rotating, and downscaling, each with a 50% probability (Figure 4). These augmentations were essential for improving the performance of all ODMs (baseline, zero-shot and few-shot models) with limited annotated samples.

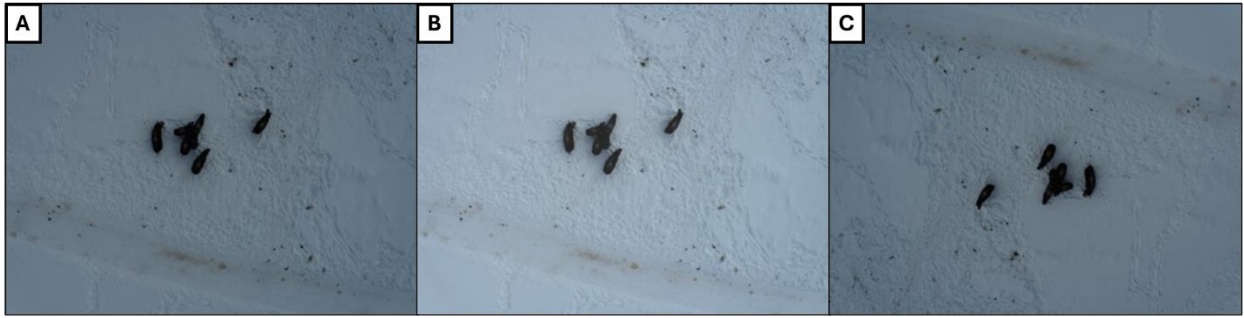

**Figure 4.** Examples of classical data augmentation applied to muskox images. The original image (A) underwent modifications with randomized brightness contrast (B) and flipping (C).

### 2.2.5. Object Detection Model Training and Optimization

The HerdNet architecture (Delplanque et al., 2023a), adapted from its original GitHub implementation, was selected for the ODM in this study. As a novel point-based object detector, HerdNet is specifically designed for accurate counting in dense groups and complex scenarios. It employs crowd counting-inspired techniques to effectively manage high-density object distributions. Tailored for detecting and counting animals, HerdNet has demonstrated superior performance compared to Faster R-CNN, making it a promising tool for wildlife monitoring. Additionally, its use of point labels significantly accelerates the image annotation process, offering a more efficient and accurate tool for detecting and counting animals in aerial imagery.

To optimize ODM performance, multiple training iterations were performed with incremental adjustments to model's weights (see Appendix S2). Each iteration was evaluated using precision, recall and F1 score metrics. Precision measures the accuracy of positive predictions, while recall measures the detection rate of all relevant instances. The F1 score, which is the harmonic mean of precision and recall, was prioritized over mean average precision (mAP), as the ODM is trained to detect a single object class (i.e., muskoxen). Bilinear interpolation was used for image resizing to reduce artifacts and distortions. Bounding box labels were converted to point labels by using the centroid of each bounding box. During patch creation, partial labels were retained if at least 50% of their surface area overlapped with the patch. Some parameters used for preprocessing the training data, such as patch width, height, overlap, number of epochs, and learning rate, remained unchanged throughout the optimization process.

To mitigate biases stemming from arbitrary random image allocation between training and validation, the best ODM underwent retraining using 5-fold cross-validation (see Appendix S3). Images from the training dataset were randomly partitioned into five equal-sized groups. Each group was then utilized once as a validation set, while the remaining groups served as training sets during each iteration.

### 2.2.6. Statistical Analysis

Statistical tests were used to evaluate differences between models' performance metrics (precision, recall, and F1 score) on the test dataset. Normality of the data was first assessed with Shapiro-Wilk test, followed by Levene test for variance homogeneity. ANOVA was used to compare means when normality was observed. Otherwise, the Kruskal Wallis test was applied to compare medians. Significant results ($p$ value $< 0.05$) led to post hoc tests: Dunn test for Kruskal-Wallis and Tukey HSD for ANOVA, revealing significant pairwise differences between models.

## 3. Results

### 3.1. Zero-Shot Object Detection Models

Increased numbers of images and patches considered in zero-shot models (Figure 5) resulted in increases precision, from $0.76 \pm 0.03$ for the ZS1 model to a value of 0.89 for the ZS4 ($\pm 0.01$) and ZS5 ($\pm 0.04$) models, along with a moderate increase in recall, from $0.76 \pm 0.02$ for ZS1 to 0.80 for ZS3 ($\pm 0.04$) and ZS4 ($\pm 0.01$). The F1 score also increased from $0.76 \pm 0.02$ for ZS1 to $0.84 \pm 0.01$ for ZS4.

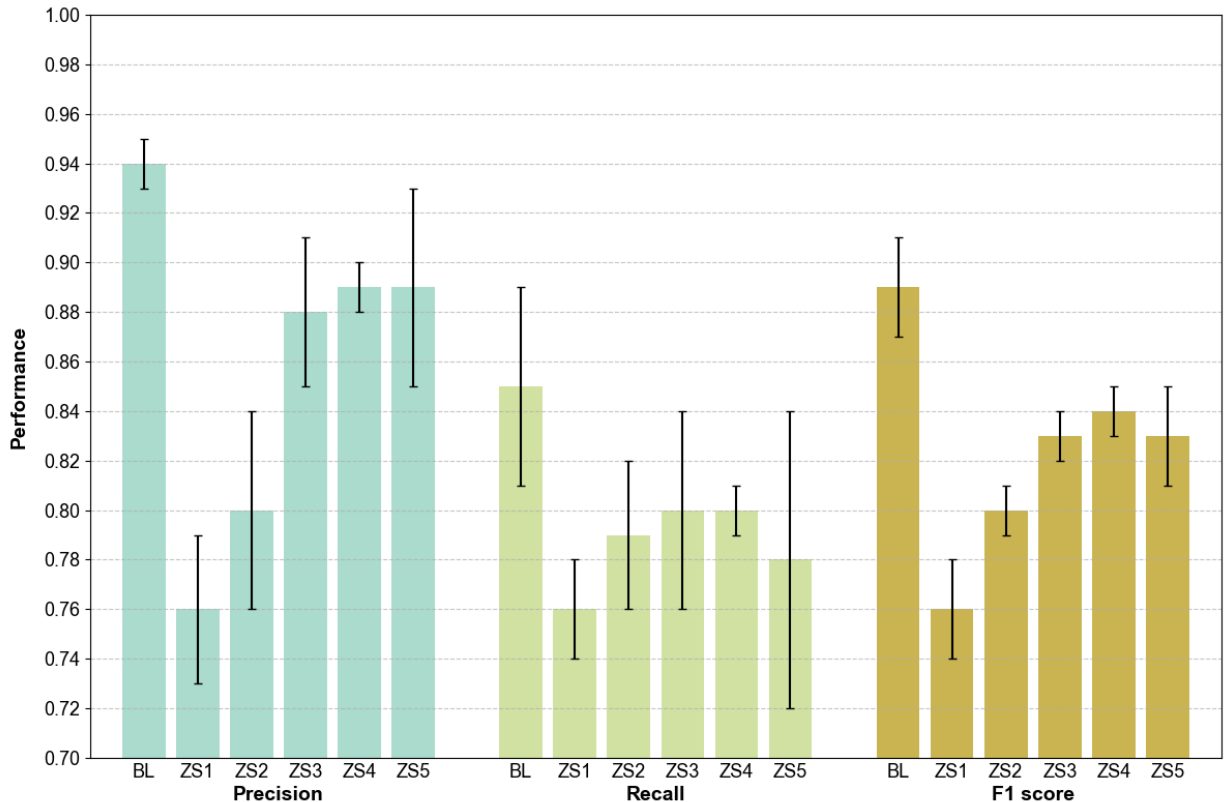

**Figure 5.** Comparison of performance metrics for the baseline (BL) and the zero-shot (ZS1-ZS5) models, including precision, recall, and F1 score. Standard deviation obtained from 5-fold cross-validation is visually represented as a vertical interval at the top of each bar.

F1 scores of all zero-shot models were significantly lower than that of the baseline model (Figure 6). When comparing zero-shot models together, F1 score significantly increased from ZS1 to ZS2 and again from ZS2 to ZS3. This reflects a progressive improvement in the model's ability to balance precision and recall as more synthetic data is added. This improvement plateaued at 96 synthetic images, with ZS4 and ZS5 showing no significant difference from ZS3.

Precision of ZS1 and ZS2 was significantly lower than the baseline model, while ZS3, ZS4, and ZS5 were not significantly different (Figure 6). Among zero-shot models, precision significantly improved from ZS1 to ZS2 and from ZS2 to ZS3. This suggests that incrementally increasing the number of synthetic images enhances the model's ability to reduce false positives. However, like F1 score, precision gains plateaued at 96 synthetic images (ZS3), as no further significant differences were observed beyond this point.

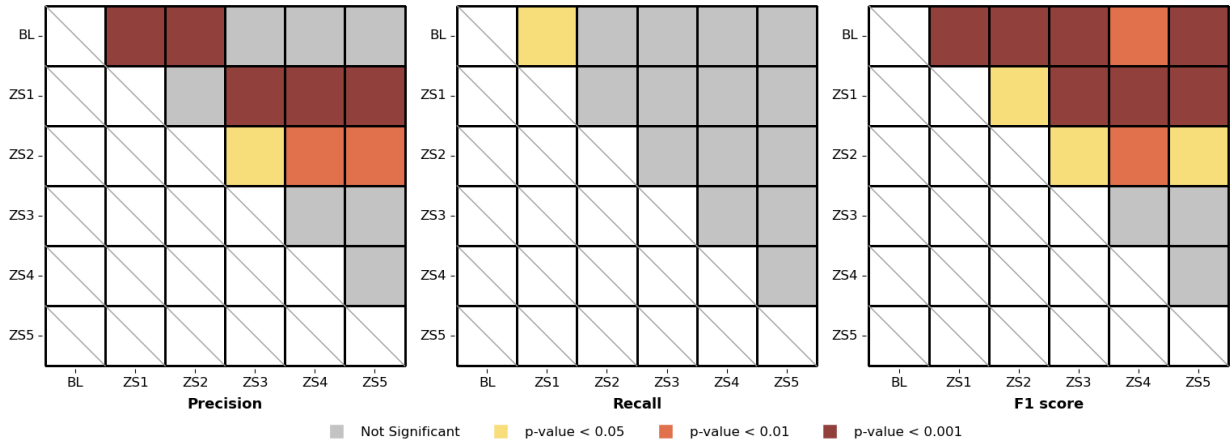

**Figure 6.** Comparative analysis of performance metrics (precision, recall and F1 score) for the baseline (BL) and the zero-shot (ZS1-ZS5) models. An ANOVA was performed on precision, recall and F1 score.

The recall value for ZS1 was significantly lower than that of the baseline model, whereas ZS2, ZS3, ZS4 and ZS5 had wider confidence intervals and were not significantly different than the baseline model. Additionally, the recall values for ZS2, ZS3, ZS4 and ZS5 were significantly different from each other (Figure 6).

Finally, adding synthetic images to zero-shot models did not have a stabilizing effect. The confidence intervals remained significantly unchanged between ZS1 and ZS5 (Figure 5), indicating persistent variability in performance.

## 3.2. Few-Shot Object Detection Models

Adding synthetic images and patches to few-shot models (Figure 7) decreased precision (about 4%) compared with the baseline model, with the lowest value of $0.90 \pm 0.01$ obtained by the FS3 model. Concurrently, there is an apparent increase in recall (about 8%) and F1 score compared to the baseline model, with performances increasing respectively up to $0.93 \pm 0.02$ for the FS3 model and up to $0.91 \pm 0.01$ for the FS4 model.

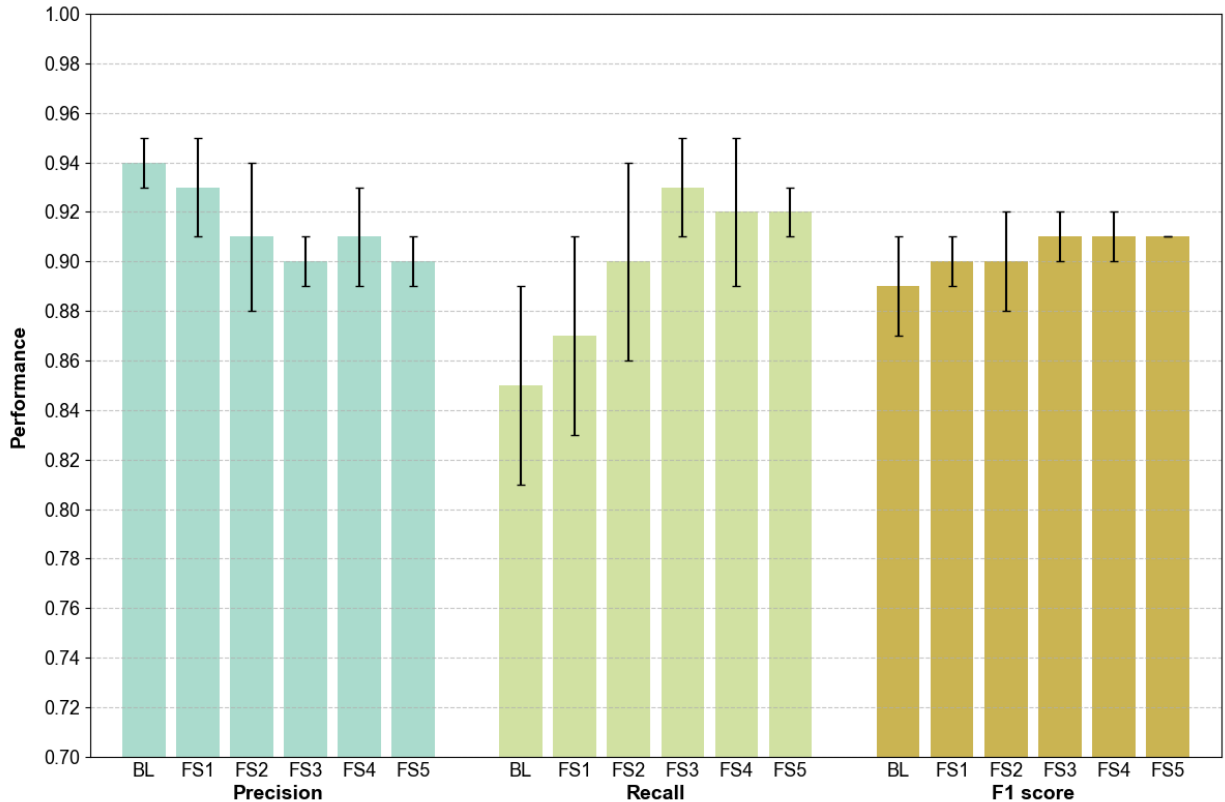

**Figure 7.** Comparison of performance metrics for the baseline (BL) and the few-shot (FS1-FS5) models, including precision, recall, and F1 score. Standard deviation obtained from 5-fold cross-validation is visually represented as a vertical interval at the top of each bar.

Precision for the FS3 and FS5 models significantly decreased compared to the baseline model (Figure 8). This indicates that adding 96 synthetic images (100% of the original dataset; FS3 model) or more to the original dataset can lead to a significant decrease in precision. In contrast, recall showed an apparent increase with the addition of synthetic images, but these differences were not significant compared to the baseline model. While FS3 had the highest recall out of all few-shot models, it was not significantly different compared to the other models. This suggests that adding synthetic data may improve recall, but the effect is not strong enough to be significant. Moreover, the differences in the F1 score between the baseline and few-shot models were not statistically significant, indicating that the balance between precision and recall remains stable despite changes in training data composition.

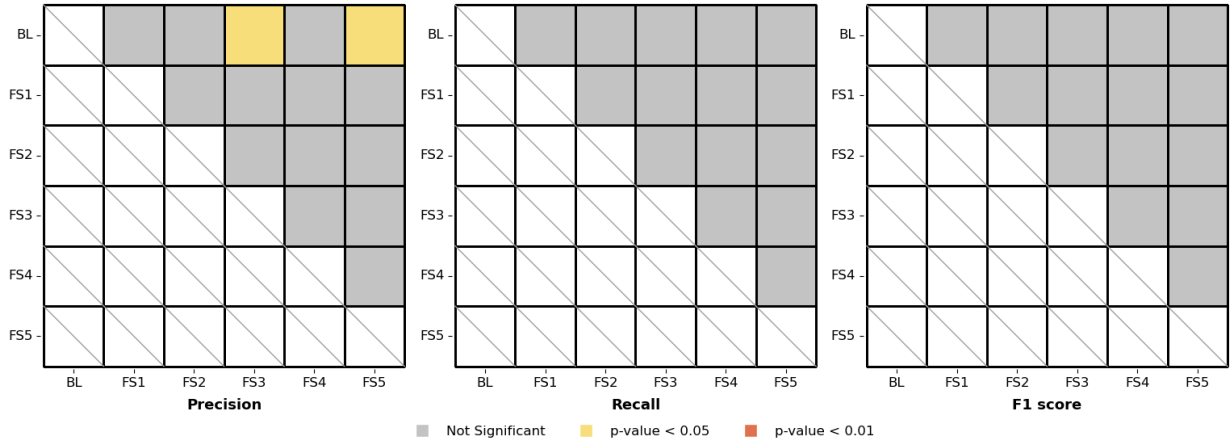

**Figure 8.** Comparative analysis of performance metrics (precision, recall and F1 score) for the baseline (BL) and the few-shot (FS1-FS5) models. An ANOVA was performed on precision, while the Kruskal-Wallis test was applied to recall and F1 score.

Finally, the addition of synthetic images in few-shot models resulted in a gradual decrease in the confidence intervals for precision, recall, and F1 score (Figure 7). Adding synthetic images therefore enhanced the model's stability and performance consistency. It is also important to note that few-shot models contain significantly more images than zero-shot models, ranging from 1.6 times as many in FS5 compared to ZS5, to 4.2 times as many in FS1 compared to ZS1 (Table 2).

### 3.3. Detection Statistics

To assess ODM behavior in a deployment-like setting, we calculated true positives, false positives, and false negatives per patch for three representative models: the baseline, zero-shot (ZS3), and few-shot (FS3) (Table 4). On average, across five folds, the FS3 model yielded the fewest missed detections (0.32), compared with the baseline (0.63) and ZS3 (0.87), indicating improved coverage. However, this came with a modest increase in false positives (0.45) relative to the baseline (0.25), while ZS3 showed the highest false positive rate (0.49). These results suggest that synthetic data, particularly in few-shot settings, can effectively reduce missed detections, albeit at the cost of a slightly greater manual review burden, defined here as the human effort required to verify and filter model predictions. In contrast, the substantially higher false positive rate of ZS3 indicates a greater post-processing workload in the absence of real training data.

**Table 4.** Detection and classification performance (true positives, false positives and false negatives) of the baseline (BL), few-shot (FS3), and zero-shot (ZS3) models (5-fold average).

| Model | Total count [1] | | | Average [2] | | |
| --- | --- | --- | --- | --- | --- | --- |
| | True positive | False positive | False negative | True positive | False positive | False negative |
| BL | 52,957 | 3,624 | 9,122 | 3.64 | 0.25 | 0.63 |
| FS3 | 57,482 | 6,534 | 4,597 | 3.96 | 0.45 | 0.32 |
| ZS3 | 49,494 | 7,081 | 12,585 | 3.41 | 0.49 | 0.87 |

[1] Total counts are calculated for all 14,533 patches, each containing one or more muskoxen.

[2] Averages are computed as total counts divided by the number of patches.

## 4.   Discussion

### 4.1.   Zero-Shot Learning: A Promising Alternative for Data Scarcity

The application of zero-shot strategies in this study yielded promising results, reaching over 80% detection of muskoxen in real images using an ODM trained exclusively on synthetic images. This result underscores the potential of zero-shot strategies as an initial approach in wildlife monitoring, particularly for developing preliminary ODMs in the absence of real-world imagery. By utilizing synthetic images and fine-tuning a pre-trained model on ImageNet, a large-scale dataset commonly used for training deep learning models, zero-shot offers a practical solution to the challenges of animal detection in data-scarce regions or for rare species with little or no available imagery. Despite the promise of synthetic data in deep learning applications, relatively few studies have explored its use in wildlife surveys. Other studies show that synthetic data can improve model generalization for underrepresented classes. Beery et al. (2020) found that synthetic examples created from video game engines boosted classification accuracy for rare species in camera trap data, while Shin et al. (2023) used generative models to balance long-tailed medical imaging datasets. These findings suggest similar gains may be possible in wildlife monitoring when real images are scarce or imbalanced.

These findings align with the work of Nguyen et al. (2024), in which a YOLOv5s model was trained on synthetic images from simulated environments for robotics and autonomous detection systems. Their model achieved a mAP of 93.1% at the 0.5:0.95 Intersection over Union threshold, further demonstrating the effectiveness of synthetic datasets in training robust ODMs. Similarly, our findings are consistent with recent research by He et al. (2022), which shows that incorporating synthetic data can significantly enhance classification results in zero-shot and few-shot recognition by improving the classifier's learning ability. However, our study also suggests that while synthetic data can be valuable, performance gains may depend on factors such as dataset size and model architecture, highlighting the need for further exploration in wildlife monitoring contexts.

Adapting and refining text prompts used for synthetic images creation also grants synthetic images a wide array of benefits. These include the capacity to simulate rare, dangerous, or challenging scenarios (e.g. monitoring wildlife in remote Arctic regions where sending manned crews poses significant logistical and safety risks), to mitigate biases in real-world datasets, and to precisely control the diversity of training conditions. These capabilities make synthetic images particularly valuable for wildlife survey efforts in inaccessible or hazardous environments.

However, the observed performance variability in zero-shot models and the diminishing returns of additional synthetic images highlight the importance of subsequent fine-tuning with real images to improve ODM generalizability and robustness. This observation is further supported by He et al. (2022), wo demonstrates that while synthetic data can greatly enhance model performance, fine-tuning with even a small number of real images can help reduce domain gaps and further improve classification accuracy, although the extent of improvement may vary depending on dataset composition and training strategy. Moreover, findings by He et al. (2022) on large-scale model pre-training suggest that synthetic data can serve as a viable substitute for real images in transfer learning, although their effectiveness may depend on the characteristics of the task and dataset.

### 4.2.   Few-Shot Learning: Balancing Real and Synthetic Data

Few-shot strategies, which use a combination of real and synthetic images, offer a practical and effective middle ground solution between ODMs trained solely on real or synthetic images. In our study, incorporating synthetic images into a real-image dataset marginally improved recall and stabilized model performance compared to the baseline model when real data was scarce. However, over-reliance on synthetic images, exceeding about twice the size of the original dataset in this study, led to observed decreases in precision (i.e., more false positives). This emphasizes the importance of carefully balancing synthetic image inclusion to optimize ODM performance.

Few-shot is particularly valuable for constructing ODMs in scenarios where real images are limited or lack diversity, as synthetic images can effectively augment real-world datasets and enhance model generalization. These findings somewhat align with research in computer vision and intelligent urban systems, where models trained using a combination of classically augmented real and synthetic data performed similarly or better than those trained solely on real or synthetic data (Abu Alhaija et al., 2018; Lin et al., 2023; Shafaei et al., 2016; Tremblay et al., 2018).

### 4.3. Role of Real Images in Training Object Detection Models

While synthetic images play an important role in supplementing data and enabling early-stage ODM development, field-acquired imagery remains essential for training wildlife monitoring ODMs. Real imagery provides ecological and contextual authenticity that are difficult to replicate synthetically. Our findings advocate a hybrid approach, strategically combining real and synthetic data to enhance model performance, especially in data-scarce settings.

### 4.4. Implications for Wildlife Monitoring Efforts

Future wildlife surveys, particularly for species showing seasonal aggregative behaviour in open habitats, should consider incorporating nadir imagery in their protocol rather than relying solely on oblique aerial imagery. Nadir views, captured directly overhead, offer distinct advantages for monitoring animals in environments like the Arctic tundra, grasslands, or other open landscapes. These advantages include reducing occlusion, improving counting accuracy in dense groups, and enabling surveys at higher altitudes, thereby minimizing disturbance to wildlife. Currently, oblique imagery is often used in muskoxen surveys to refine visually interpreted counts and assist in determining sex and age (Davison and Williams, 2022). While oblique imagery provides valuable benefits, such as enhanced visibility of distinguishing features (Adamczewski et al., 2021) and the ability to reveal animals partially obscured by vegetation (Lamprey et al., 2020), it can also introduce challenges like occlusion in large groups. Given these factors, nadir imagery represents a more practical approach for accurately assessing population sizes and group structure in species that inhabit open landscapes, including but not limited to muskoxen.

The use of ODMs, such as HerdNet, can streamline the traditionally labor-intensive process of manual image review. Typically, biologists must manually count animals from images to verify population estimates (Adamczewski et al., 2021), but an ODM could automatically generate preliminary labels, allowing experts to focus on refining these annotations rather than conducting the entire count from scratch. Such an approach would reduce manual workload, observer fatigue, and interpretation bias, while also enhancing the objectivity and consistency of results (Delplanque et al., 2024a, 2023b).

Moreover, the use of synthetic images can extend beyond muskoxen to other species, offering substantial benefits in any context where aerial imagery is used to survey wildlife populations. It could prove useful for species with limited available imagery, allowing models to be initialized for species recognition even when only a small dataset or no data at all is available. Over time, as more real images of a species are collected through surveys and made available, the need for synthetic images may decrease. Therefore, data augmentation with synthetic images is especially valuable during the initial phases of monitoring new or understudied species.

While this study assessed the model's performance using metrics like precision, recall, and F1 score on real images of muskoxen, future work could focus on deriving population estimates directly from ODM detections. These abundance estimates could then be compared to those obtained through traditional wildlife surveys, providing valuable insights into potential biases and discrepancies. Notably, previous work has shown that counts from aerial imagery can be more precise than observer-based counts (Hodgson et al., 2018), supporting a shift toward image-based pipelines. Such comparisons would not only help quantify the ODM's detection probabilities in relation to those from aerial transect sampling, but also validate its accuracy and reliability for population monitoring. Continued testing under real-world conditions could further demonstrate the model's applicability, reinforcing its role as a complementary or alternative tool for wildlife management and conservation decision-making.

### 4.5.   Limitations

A notable limitation in using synthetic images for training ODMs is the potential introduction of artifacts or patterns absent from real-world data distributions. This challenge is particularly pronounced in multi-species contexts, where distinguishing between target species demands high precision (Delplanque et al., 2023b). Synthetic images created by multimedia software or generative models often differ from real imagery due to variations in the image formation process, resulting in distinct inherent characteristics (Corvi et al., 2023; Meena and Tyagi, 2021, 2019; Verdoliva, 2020). Additionally, synthetic images can display unnatural qualities (Chen et al., 2023; Nirkin et al., 2022; Peng et al., 2024), such as disproportionate animals, blurriness around object edges, or unrealistic ecological contexts where animals appear fused or arranged unnaturally within a group. This unrealism could further exacerbate detection errors. Cut-and-paste methods can also be used to preserve object and background realism. However, tests performed using this approach in our study showed that while model performance can be similar to that obtained with DALL-E 2, it requires considerably more manual effort and relies on background image diversity that is rarely available in the context of FSL and ZSL. While synthetic images enhance generalization by increasing training diversity, over-reliance can lead to overfitting (Abu Alhaija et al., 2018), where the ODM misinterprets synthetic features as real, increasing false positives and reducing precision.

Another limitation lies in the challenge of isolating the specific contributions of synthetic image quality versus data volume, especially in low-data settings. While performance improvements were observed when adding synthetic data, further research is required to determine whether these gains are attributable to the diversity of the synthetic imagery or simply to the increased size of the training set.

Diffusion-based generative models, such as DALL-E 2, represent a promising avenue for generating highly realistic synthetic images by progressively transforming random noise into coherent images. These models show promise for generating highly realistic images of animal species, which can bootstrap automated processes for detection and counting. However, they also face several limitations. One key issue encountered in this study was the difficulty of generating contextually accurate imagery for niche subjects like muskoxen. When specific keywords such as "muskox" and "nadir" were combined in prompts, DALL-E 2 often produced mismatched perspectives, with nadir environments but oblique animal views (Figure 9). This mismatch likely stems from the lack of diverse nadir-view muskox images in the training data, which predominantly features oblique angles. To our knowledge, nadir images, often captured during photocensus studies, are not publicly accessible. The issue may also be exacerbated by the model's difficulty in fully interpreting nuanced language in prompts, leading to images that fail to fully reflect the intended meaning or context (Leivada et al., 2023).

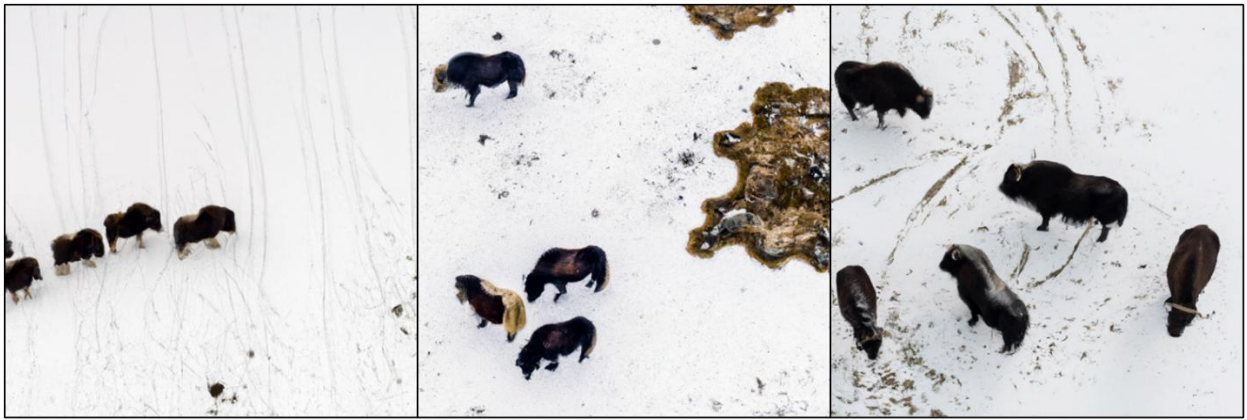

**Figure 9.** Examples of discarded synthetic images from DALL-E 2 exhibiting perspective mismatch when combining "*muskox*" and "*nadir*" keywords.

Attempts to enrich the dataset by searching 20 image-sharing platforms (e.g., Dronestagram, Flickr, Imgur) for additional nadir-like muskox images yielded no results, reinforcing the hypothesis that DALL-E 2's training data is skewed towards more common oblique views. Furthermore, the model often struggled to render realistic depictions of muskoxen, producing images with blended features, distorted anatomy, or unnatural postures (Figure 10). Such inaccuracies complicated the annotation process, as distorted depictions hindered clear identification and labeling of subjects. Consequently, synthetic images were manually screened for quality, with 84% of generated images discarded due to unrealistic outputs. Despite the high rejection rate, filtering was fast and far less resource-intensive than acquiring and labeling real imagery.

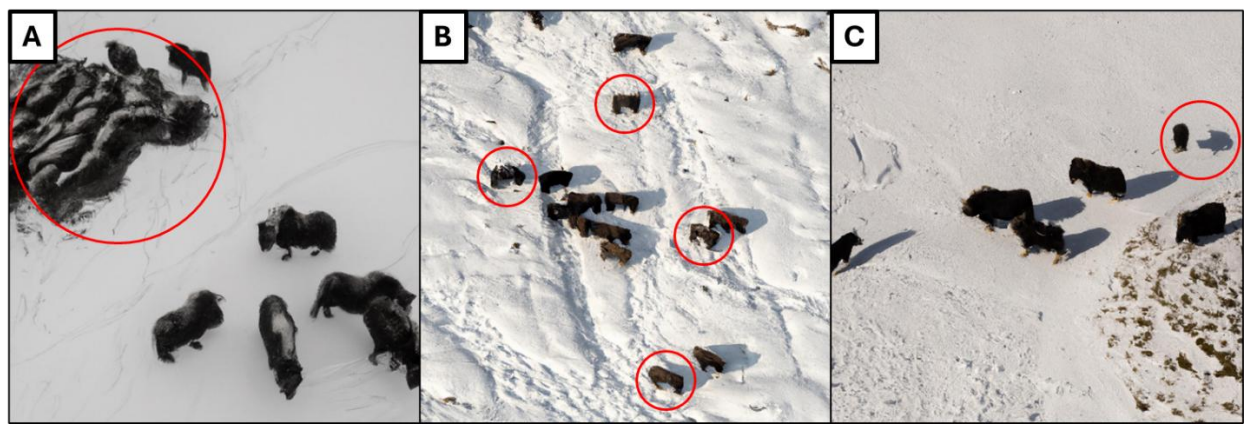

**Figure 10.** Examples of discarded synthetic images from DALL-E 2 exhibiting unrealistic depictions of muskoxen. A) Muskoxen appearing as an indistinct blob. B) Confusion between muskoxen and rocks. C) A partial muskox with a shadow detached from its body.

Another limitation of DALL-E 2 lies in its fixed output resolution of 256 x 256, 512 x 512, or 1024 x 1024 pixels, which are smaller than the high-resolution aerial images used in this study. Resizing was therefore applied to standardize muskox sizes between real and synthetic images. Despite being much more cost-effective than field surveys, DALL-E 2 incurs a cost of $0.02 per

image at maximum resolution. Furthermore, batch generation is limited to 10 images at a time, which constrained generation scalability. These limitations may be alleviated with future improvements in generative models.

### 4.6. Future Research Directions

### 4.6.1. Next-Generation of Synthetic Images

Future research should explore the impact of advanced techniques for synthetic image creation and their impact on the performance of ODMs. Special emphasis should be given to modern generative models for image synthesis, including likelihood-based diffusion models, currently recognized as state-of-the-art on benchmarks such as CIFAR-10, and generative adversarial networks, which remain widely used and relevant for domain-specific tasks and controlled testing datasets like LSUN and ImageNet (Dhariwal and Nichol, 2021; Mehmood et al., 2024; C. Zhang et al., 2023). The rapid advancement of diffusion-based models highlights the importance of reassessing model selection. DALL-E 2 was chosen in this study for its then leading edge in visual quality, particularly in rendering accurate lighting, shadows, and textures, as well as its unique editing capabilities such as inpainting and outpainting. Since then, numerous other models have emerged, including DALL-E 3 (OpenAI, 2023), which builds on its predecessor's strengths, alongside popular alternatives like Stable Diffusion (Stability AI, 2024), Midjourney (Midjourney, 2023) and Imagen 3 (Imagen 3 Team, Google, 2024). However, complementary analysis has shown that recent models, such as Gemini (Gemini 2.5 Flash Image), are unable to produce realistic images in the context of a wildlife survey. Other models, such as DALL-E 3, have shown better realism in the images produced in this context compared to DALL-E 2, but these improvements have not translated into a clear increase in model performance (Appendix S4). As generative models continue to improve, the usability of synthetic images is expected to increase, reducing the need for manual filtering and further supporting their adoption in wildlife applications.

Further exploration is also needed in more complex ecological scenarios, such as multi-species detection and cluttered natural environments, to test ODM generalizability. Additionally, assessing how the varying characteristics of synthetic images (e.g., resolution, noise levels, feature diversity) influence detection outcomes could refine synthetic data strategies. Combining outputs from multiple generative models or overlaying synthetic objects onto real backgrounds, as done in (Abu Alhaija et al., 2018), may also enhance the robustness and applicability of ODMs trained with synthetic data.

Additionally, prompt engineering, the practice of designing and refining input text prompts for AI models, represents a significant area for future research in the creation of synthetic images using diffusion models. Research shows that this process is essential for enhancing the quality and relevance of synthetic images, as the formulation of prompts can dramatically affect the resulting output (Giray, 2023; Hao et al., 2022; Wang et al., 2023; Zhou et al., 2022).

### 4.6.2. Real and Synthetic Satellite Images

Interest in using very high resolution (VHR) satellite imagery for wildlife applications, particularly for animal population surveys, is rapidly growing (Delplanque et al., 2024b). Improvements in spatial and temporal resolutions make satellite imagery a promising tool for efficiently surveying large areas, including remote or dangerous locations, while minimizing wildlife disturbance.

However, integrating VHR satellite imagery into ODM datasets poses significant challenges. The limited availability and high cost of satellite imagery restrict both the spatial coverage and the frequency of acquisitions. Furthermore, coordinating satellite passes with animal movements remains one of the most significant obstacles to an effective use, as animals may shift locations between image captures (Delplanque et al., 2024b). Careful consideration is needed of how VHR imagery is used, particularly when incorporating it into workflows that rely on generative models, which require consistent and high-quality input data to perform effectively.

Until VHR satellite images become more widely available, feasibility studies using synthetic satellite images could offer valuable insights. Future research should explore the performance of ODMs trained on synthetic satellite imagery, assessing their effectiveness in detecting animals across diverse landscapes and image resolutions compared to traditional aerial imagery. To maximize the utility of this approach, the development of specialized generative models tailored for wildlife applications is essential. Emerging advancements in domain-specific generative models (Khanna et al., 2023) could offer the best quality and realism for synthetic satellite imagery. These innovations could bridge the gap in data availability and expand the application of generative models in wildlife monitoring.

### Acknowledgements

The work of Simon DURAND was supported by the Natural Sciences and Engineering Research Council of Canada (NSERC) and the Fonds de Recherche du Québec Nature and Technologies (FRQNT). Special thanks are extended to individuals and organizations involved in the acquisition of data: Alexandre Paiement, Vincent Brodeur and William Rondeau from the *Ministère de l'Environnement, de la Lutte contre les Changements Climatiques, de la Faune et des Parcs* from the Government of Quebec, the Department of Environment and Natural Resources from the Government of the Northwest Territories, David Pagé from the Zoo Sauvage de Saint-Félicien, Marc-André Bureau from the Cégep de Saint-Félicien, Matt from Etherium Sky, and Mark Austin from the Musk Ox Farm.

### Author's Contributions Statement

Simon Durand, Samuel Foucher and Jérôme Théau conceived the ideas and designed the methodology; Simon Durand, Jérôme Théau and Joëlle Taillon collected the data; Simon Durand, Samuel Foucher, Alexandre Delplanque and Jérôme Théau analysed the data; Simon Durand led the writing of the manuscript; and Simon Durand, Samuel Foucher, Alexandre Delplanque, Joëlle Taillon and Jérôme Théau contributed critically to the drafts. All authors gave final approval for publication.

### Author's Declaration

All authors have reviewed and approved the final version of the manuscript. Each author has made significant contributions to the work, and all individuals meeting the criteria for authorship have been included.

### Data sharing

Our image data (synthetic images and Zoo Sauvage de Saint-Félicien), annotations, and Python code are available at https://doi.org/10.5683/SP3/LCOG2G. The HerdNet architecture used in this study is available as open-source code on GitHub at https://github.com/Alexandre-Delplanque/HerdNet.

## References

Abu Alhaija, H., Mustikovela, S.K., Mescheder, L., Geiger, A., Rother, C., 2018. Augmented reality meets computer vision: efficient data generation for urban driving scenes. International Journal of Computer Vision 126(9), 961–972. https://doi.org/10.1007/s11263-018-1070-x

Adamczewski, J., Olesen, K., Olesen, D., Williams, J., Cluff, D., Boulanger, J., 2021. Late winter 2018 muskox photo composition survey, East Arm of Great Slave Lake (Manuscript Report No. 296). Government of Northwest Territories, Environment and Natural Resources, pp. 41.

Alaska Department of Fish and Game, 2001. Muskox management report of survey-inventory activities 1 July 1998-30 June 2000 (Management Report). Juneau, Alaska, United States of America, pp. 55.

AMAP, 2017. Snow, Water, Ice and Permafrost in the Arctic (SWIPA) 2017. Arctic Monitoring and Assessment Programme (AMAP), Oslo, Norway, pp. xiv + 269.

Anderson, M., 2016. Distribution and abundance of muskoxen (Ovibos moschatus) and Peary caribou (Rangifer tarandus pearyi) on Prince of Wales, Somerset, and Russell Islands, August 2016 (Status Report No. 2016– 06). Government of Nunavut, Department of Environment, Wildlife Research Section, Igloolik, Nunavut, Canada, pp. v + 22.

Antonelli, S., Avola, D., Cinque, L., Crisostomi, D., Foresti, G.L., Galasso, F., Marini, M.R., Mecca, A., Pannone, D., 2022. Few-shot object detection: a survey. ACM Computing Surveys 54(11s), 37. https://doi.org/10.1145/3519022

Beery, S., Liu, Y., Morris, D., Piavis, J., Kapoor, A., Meister, M., Joshi, N., Perona, P., 2020. Synthetic examples improve generalization for rare classes, in: 2020 IEEE Winter Conference on Applications of Computer Vision (WACV). Presented at the 2020 IEEE Winter Conference on Applications of Computer Vision (WACV), IEEE, Snowmass Village, CO, USA, pp. 852–862. https://doi.org/10.1109/WACV45572.2020.9093570

Brodeur, A., Leblond, M., Brodeur, V., Taillon, J., Côté, S.D., 2023. Investigating potential for competition between migratory caribou and introduced muskoxen. The Journal of Wildlife Management 87(3), e22366, 24. https://doi.org/10.1002/jwmg.22366

Buckland, S.T., Anderson, D.R., Burnham, K.P., Laake, J.L., Borchers, D.L., Thomas, L., 2001. Introduction to distance sampling: estimating abundance of biological populations, 1st ed. Oxford University Press, Oxford, Toronto, Canada, pp. 448.

Buslaev, A., Iglovikov, V.I., Khvedchenya, E., Parinov, A., Druzhinin, M., Kalinin, A.A., 2020. Albumentations: fast and flexible image augmentations. Information 11(2), 125. https://doi.org/10.3390/info11020125

Caughley, G., 1974. Bias in aerial survey. The Journal of Wildlife Management 38(4), 921. https://doi.org/10.2307/3800067

Chen, Z., Sun, W., Wu, H., Zhang, Z., Jia, J., Ji, Z., Sun, F., Jui, S., Min, X., Zhai, G., Zhang, W., 2023. Exploring the naturalness of AI-generated images. arXiv preprint, pp. 33. https://doi.org/10.48550/ARXIV.2312.05476

Corvi, R., Cozzolino, D., Zingarini, G., Poggi, G., Nagano, K., Verdoliva, L., 2023. On the detection of synthetic images generated by diffusion models, in: ICASSP 2023 - 2023 IEEE International Conference on Acoustics, Speech and Signal Processing (ICASSP). Presented at the ICASSP 2023 - 2023 IEEE International Conference on Acoustics, Speech and Signal Processing (ICASSP), IEEE, Rhodes Island, Greece, p. 5. https://doi.org/10.1109/ICASSP49357.2023.10095167

Couturier, S., Dale, A., Wood, B., Snook, J., 2018. Results of a spring 2017 aerial survey of the Torngat Mountains Caribou Herd. Torngat Wildlife, Plants and Fisheries Secretariat, Happy Valley-Goose Bay, Newfoundland and Labrador, Canada. pp. xiv + 50.

Cuyler, C., Rowell, J., Adamczewski, J., Anderson, M., Blake, J., Bretten, T., Brodeur, V., Campbell, M., Checkley, S.L., Cluff, H.D., Côté, S.D., Davison, T., Dumond, M., Ford, B., Gruzdev, A., Gunn, A., Jones, P., Kutz, S., Leclerc, L.-M., Mallory, C., Mavrot, F., Mosbacher, J.B., Okhlopkov, I.M., Reynolds, P., Schmidt, N.M., Sipko, T., Suitor, M., Tomaselli, M., Ytrehus, B., 2020. Muskox status, recent variation, and uncertain future. Ambio 49(3), 805–819. https://doi.org/10.1007/s13280-019-01205-x

Davison, T., Williams, J., 2022. Aerial survey of muskoxen (Ovibos moschatus) and Peary caribou (Rangifer tarandus pearyi) on Northwest Victoria Island, May 2019 (Manuscript Report No. 303). Government of Northwest Territories, Environment and Natural Resources, pp. 17.

Delplanque, A., Foucher, S., Théau, J., Bussière, E., Vermeulen, C., Lejeune, P., 2023a. From crowd to herd counting: how to precisely detect and count African mammals using aerial imagery and deep learning? ISPRS Journal of Photogrammetry and Remote Sensing 197, 167–180. https://doi.org/10.1016/j.isprsjprs.2023.01.025

Delplanque, A., Lamprey, R., Foucher, S., Théau, J., Lejeune, P., 2023b. Surveying wildlife and livestock in Uganda with aerial cameras: deep Learning reduces the workload of human interpretation by over 70%. Frontiers in Ecology and Evolution 11, 9. https://doi.org/10.3389/fevo.2023.1270857

Delplanque, A., Linchant, J., Vincke, X., Lamprey, R., Théau, J., Vermeulen, C., Foucher, S., Ouattara, A., Kouadio, R., Lejeune, P., 2024a. Will artificial intelligence revolutionize aerial surveys? A first large-scale semi-automated survey of African wildlife using oblique imagery and deep learning. Ecological Informatics 82, 10. https://doi.org/10.1016/j.ecoinf.2024.102679

Delplanque, A., Théau, J., Foucher, S., Serati, G., Durand, S., Lejeune, P., 2024b. Wildlife detection, counting and survey using satellite imagery: are we there yet? GIScience & Remote Sensing 61(1), 30. https://doi.org/10.1080/15481603.2024.2348863

Dhariwal, P., Nichol, A., 2021. Diffusion models beat GANs on image synthesis. arXiv preprint, pp. 44. https://doi.org/10.48550/ARXIV.2105.05233

Eikelboom, J.A.J., Wind, J., Van De Ven, E., Kenana, L.M., Schroder, B., De Knegt, H.J., Van Langevelde, F., Prins, H.H.T., 2019. Improving the precision and accuracy of animal population estimates with aerial image object detection. Methods in Ecology and Evolution 10(11), 1875–1887. https://doi.org/10.1111/2041-210X.13277

Environment and Climate Change Canada, 2022. Guidance and protocols for wildlife surveys for emergency response. Gatineau, Quebec, Canada, pp. x + 97.

Fleming, P.J.S., Tracey, J.P., 2008. Some human, aircraft and animal factors affecting aerial surveys: how to enumerate animals from the air. Wildlife Research 35(4), 258. https://doi.org/10.1071/WR07081

Giray, L., 2023. Prompt engineering with ChatGPT: a guide for academic writers. Annals of Biomedical Engineering 51(12), 2629–2633. https://doi.org/10.1007/s10439-023-03272-4

Gonzalez, L.F., Montes, G.A., Puig, E., Johnson, S., Mengersen, K., Gaston, K.J., 2016. Unmanned aerial vehicles (UAVs) and artificial intelligence revolutionizing wildlife monitoring and conservation. Sensors 16(1), 97. https://doi.org/10.3390/s16010097

Gunn, A., Adamczewski, J., 2003. Muskox: Ovibos moschatus, in: Wild Mammals of North America: Biology, Management, and Conservation. Johns Hopkins University Press, Baltimore, Maryland, United States of America, pp. 1076–1094.

Hao, Y., Chi, Z., Dong, L., Wei, F., 2022. Optimizing prompts for text-to-image generation. arXiv preprint, pp. 16. https://doi.org/10.48550/ARXIV.2212.09611

He, R., Sun, S., Yu, X., Xue, C., Zhang, W., Torr, P., Bai, S., Qi, X., 2022. Is synthetic data from generative models ready for image recognition? arXiv preprint, pp. 24. https://doi.org/10.48550/ARXIV.2210.07574

Hodgson, J.C., Mott, R., Baylis, S.M., Pham, T.T., Wotherspoon, S., Kilpatrick, A.D., Raja Segaran, R., Reid, I., Terauds, A., Koh, L.P., 2018. Drones count wildlife more accurately and precisely than humans. Methods in Ecology and Evolution 9(5), 1160–1167. https://doi.org/10.1111/2041-210x.12974

Huang, G., Laradji, I., Vazquez, D., Lacoste-Julien, S., Rodriguez, P., 2023. A survey of self-supervised and few-shot object detection. IEEE Transactions on Pattern Analysis and Machine Intelligence 45(4), 4071–4089. https://doi.org/10.1109/TPAMI.2022.3199617

Imagen 3 Team, Google, 2024. Imagen 3. arXiv preprint, pp. 35. https://doi.org/10.48550/ARXIV.2408.07009

Khanna, S., Liu, P., Zhou, L., Meng, C., Rombach, R., Burke, M., Lobell, D., Ermon, S., 2023. DiffusionSat: a generative foundation model for satellite imagery. arXiv preprint, pp. 19. https://doi.org/10.48550/ARXIV.2312.03606

Kirillov, A., Mintun, E., Ravi, N., Mao, H., Rolland, C., Gustafson, L., Xiao, T., Whitehead, S., Berg, A.C., Lo, W.-Y., Dollár, P., Girshick, R., 2023. Segment Anything. arXiv preprint, pp. 30. https://doi.org/10.48550/ARXIV.2304.02643

Kutz, S., Rowell, J., Adamczewski, J., Gunn, A., Cuyler, C., Aleuy, O.A., Austin, M., Berger, J., Blake, J., Bondo, K., Dalton, C., Dobson, A., Di Francesco, J., Gerlach, C., Kafle, P., Mavrot, F., Mosbacher, J., Murray, M., Nascou, A., Orsel, K., Rossouw, F., Schmidt, N.-M., Suitor, M., Tomaselli, M., Ytrehus, B., 2017. Muskox health ecology symposium 2016: gathering to share knowledge on Umingmak in a time of rapid change. Arctic 70(2), 225–236. https://doi.org/10.14430/arctic4656

Lamprey, R., Pope, F., Ngene, S., Norton-Griffiths, M., Frederick, H., Okita-Ouma, B., Douglas-Hamilton, I., 2020. Comparing an automated high-definition oblique camera system to rear-seat-observers in a wildlife survey in Tsavo, Kenya: taking multi-species aerial counts to the next level. Biological Conservation 241, 15. https://doi.org/10.1016/j.biocon.2019.108243

Lancia, R.A., Kendall, W.L., Pollock, K.H., Nichols, J.D., 2005. Estimating the number of animals in wildlife populations, in: Research and Management Techniques for Wildlife and Habitats. The Wildlife Society, Bethesda, Maryland, United States of America, pp. 106–153.

Le Moullec, M., Pedersen, Å.Ø., Yoccoz, N.G., Aanes, R., Tufto, J., Hansen, B.B., 2017. Ungulate population monitoring in an open tundra landscape: distance sampling versus total counts. Wildlife Biology 2017(1), 1–11. https://doi.org/10.2981/wlb.00299

Leboeuf, A., Morneau, C., Robitaille, A., Dufour, E., Grondin, P., 2018. Ecological mapping of the vegetation of northern Québec – Mapping standard. Direction des inventaires forestiers, Ministère des Forêts, de la Faune et des Parcs, Quebec, Canada. pp. III + 17.

Leivada, E., Murphy, E., Marcus, G., 2023. DALL·E 2 fails to reliably capture common syntactic processes. Social Sciences & Humanities Open 8, 10. https://doi.org/10.1016/j.ssaho.2023.100604

Lemay, M., Provencher-Nolet, L., Bernier, M., Lévesque, E., Boudreau, S., 2018. Spatially explicit modeling and prediction of shrub cover increase near Umiujaq, Nunavik. Ecological Monographs 88(3), 385–407. https://doi.org/10.1002/ecm.1296

Lin, S., Wang, K., Zeng, X., Zhao, R., 2023. Explore the power of synthetic data on few-shot object detection, in: 2023 IEEE/CVF Conference on Computer Vision and Pattern Recognition Workshops (CVPRW). Presented at the 2023 IEEE/CVF Conference on Computer Vision and Pattern Recognition Workshops (CVPRW), IEEE, Vancouver, British Columbia, Canada, pp. 638–647. https://doi.org/10.1109/CVPRW59228.2023.00071

Linchant, J., Lisein, J., Semeki, J., Lejeune, P., Vermeulen, C., 2015. Are unmanned aircraft systems (UASs) the future of wildlife monitoring? A review of accomplishments and challenges. Mammal Review 45(4), 239–252. https://doi.org/10.1111/mam.12046

McComb, B., Zuckerberg, B., Vesely, D., Jordan, C., 2010. Monitoring animal populations and their habitats: a practitioner's guide, 1st ed. CRC Press, Boca Raton, Florida, United States of America, pp. xi + 452. https://doi.org/10.1201/9781420070583

Meena, K.B., Tyagi, V., 2021. Distinguishing computer-generated images from photographic images using two-stream convolutional neural network. Applied Soft Computing 100, 10. https://doi.org/10.1016/j.asoc.2020.107025

Meena, K.B., Tyagi, V., 2019. Methods to distinguish photorealistic computer generated images from photographic images: a review, in: Singh, M., Gupta, P.K., Tyagi, V., Flusser, J., Ören, T., Kashyap, R. (Eds.), Advances in Computing and Data Sciences. ICACDS 2019, Communications in Computer and Information Science. Springer, Singapore, pp. 64–82. https://doi.org/10.1007/978-981-13-9939-8_7

Mehmood, R., Bashir, R., Giri, K.J., 2024. Text conditioned generative adversarial networks generating images and videos: a critical review. SN Computer Science 5(935), 30. https://doi.org/10.1007/s42979-024-03289-z

Midjourney, 2023. Midjourney [WWW Document]. URL https://www.midjourney.com/home (accessed 10.6.24).

Newey, S., Davidson, P., Nazir, S., Fairhurst, G., Verdicchio, F., Irvine, R.J., Van Der Wal, R., 2015. Limitations of recreational camera traps for wildlife management and conservation research: a practitioner's perspective. Ambio 44(S4), 624–635. https://doi.org/10.1007/s13280-015-0713-1

Nguyen, H.L., Le, D.T., Hoang, H.H., 2024. Application of synthetic data on object detection tasks. Engineering, Technology & Applied Science Research 14(4), 15695–15699. https://doi.org/10.48084/etasr.7929

Nirkin, Y., Wolf, L., Keller, Y., Hassner, T., 2022. Deepfake detection based on discrepancies between faces and their context. IEEE Transactions on Pattern Analysis and Machine Intelligence 44(10), 6111–6121. https://doi.org/10.1109/TPAMI.2021.3093446

OpenAI, 2023. DALL-E 3 [WWW Document]. URL https://openai.com/index/dall-e-3/ (accessed 10.6.24).

Peng, Q., Lu, Y., Peng, Y., Qian, S., Liu, X., Shen, C., 2024. Crafting synthetic realities: examining visual realism and misinformation potential of photorealistic AI-generated images. arXiv preprint, pp. 13. https://doi.org/10.48550/ARXIV.2409.17484

Piper, L., 2016. Great Slave Lake [WWW Document]. The Canadian Encyclopedia. URL https://www.thecanadianencyclopedia.ca/fr/article/grand-lac-des-esclaves (accessed 5.15.22).

Prosekov, A., Kuznetsov, A., Rada, A., Ivanova, S., 2020. Methods for monitoring large terrestrial animals in the wild. Forests 11(8), 808, 12. https://doi.org/10.3390/f11080808

Ramesh, A., Dhariwal, P., Nichol, A., Chu, C., Chen, M., 2022. Hierarchical text-conditional image generation with CLIP latents. arXiv preprint, pp. 27. https://doi.org/10.48550/ARXIV.2204.06125

Schlossberg, S., Chase, M.J., Griffin, C.R., 2016. Testing the accuracy of aerial surveys for large mammals: an experiment with African savanna elephants (Loxodonta africana). PLoS ONE 11(10), 19. https://doi.org/10.1371/journal.pone.0164904

Schmidt, J.H., Thompson, W.L., Wilson, T.L., Reynolds, J.H., 2022. Distance sampling surveys: using components of detection and total error to select among approaches. Wildlife Monographs 210(1), 56. https://doi.org/10.1002/wmon.1070

Shafaei, A., Little, J., Schmidt, M., 2016. Play and learn: using video games to train computer vision models, in: Procedings of the British Machine Vision Conference 2016. Presented at the British Machine Vision Conference 2016, British Machine Vision Association, York, UK, p. 18. https://doi.org/10.5244/C.30.26

Shin, J., Kang, M., Park, J., 2023. Fill-up: balancing long-tailed data with generative models. arXiv preprint, pp. 32. https://doi.org/10.48550/ARXIV.2306.07200

Sinclair, A.R.E., Fryxell, J.M., Caughley, G., 2006. Wildlife ecology, conservation, and management, 2nd ed. Blackwell Publishing, Malden, Massachusetts, United States of America, pp. xii + 469.

Stability AI, 2024. Stable Diffusion [WWW Document]. URL https://stability.ai/stable-image (accessed 10.6.24).

Tkachenko, M., Malyuk, M., Holmanyuk, A., Liubimov, N., 2020. Label Studio: data labeling software.

Tremblay, J., Prakash, A., Acuna, D., Brophy, M., Jampani, V., Anil, C., To, T., Cameracci, E., Boochoon, S., Birchfield, S., 2018. Training deep networks with synthetic data: bridging the reality gap by domain randomization, in: 2018 IEEE/CVF Conference on Computer Vision and Pattern Recognition Workshops (CVPRW). Presented at the 2018 IEEE/CVF Conference on Computer Vision and Pattern Recognition Workshops (CVPRW), IEEE Computer Society, Salt Lake City, Utah, United States of America, pp. 1082–10828. https://doi.org/10.1109/CVPRW.2018.00143

Verdoliva, L., 2020. Media forensics and deepfakes: an overview. IEEE Journal of Selected Topics in Signal Processing 14(5), 910–932. https://doi.org/10.1109/JSTSP.2020.3002101

Walters, C.J., 1986. Adaptive management of renewable resources, Biological resource management. Macmillan Publishing Company, New York, New York, United States of America, pp. 374.

Wang, D., Shao, Q., Yue, H., 2019. Surveying wild animals from satellites, manned aircraft and unmanned aerial systems (UASs): a review. Remote Sensing 11(11), 1308, 28. https://doi.org/10.3390/rs11111308

Wang, J., Liu, Z., Zhao, L., Wu, Z., Ma, C., Yu, S., Dai, H., Yang, Q., Liu, Y., Zhang, Songyao, Shi, E., Pan, Y., Zhang, T., Zhu, D., Li, X., Jiang, X., Ge, B., Yuan, Y., Shen, D., Liu, T., Zhang, Shu, 2023. Review of large vision models and visual prompt engineering. Meta-Radiology 1(3), 100047, 36. https://doi.org/10.1016/j.metrad.2023.100047

Wang, Y., Yao, Q., Kwok, J.T., Ni, L.M., 2021. Generalizing from a few examples: a survey on few-shot learning. ACM Computing Surveys 53(3), 34. https://doi.org/10.1145/3386252

Williams, B.K., Nichols, J.D., Conroy, M.J., 2002. Analysis and management of animal populations: modeling, estimation, and decision making, 1st ed. Academic Press, San Diego, California, United States of America, pp. 817.

Witmer, G.W., 2005. Wildlife population monitoring: some practical considerations. Wildlife Research, USDA National Wildlife Research Center - Staff Publications 32(3), 259–263. https://doi.org/10.1071/WR04003

Yang, J., Guo, X., Li, Y., Marinello, F., Ercisli, S., Zhang, Z., 2022. A survey of few-shot learning in smart agriculture: developments, applications, and challenges. Plant Methods 18(28), 12. https://doi.org/10.1186/s13007-022-00866-2

Zhang, C., Zhang, Chaoning, Zhang, M., Kweon, I.S., 2023. Text-to-image diffusion models in generative AI: a survey. arXiv preprint, pp. 13. https://doi.org/10.48550/ARXIV.2303.07909

Zhang, Q., Yi, X., Guo, J., Tang, Y., Feng, T., Liu, R., 2023. A few-shot rare wildlife image classification method based on style migration data augmentation. Ecological Informatics 77, 102237, 12. https://doi.org/10.1016/j.ecoinf.2023.102237

Zhou, Y., Muresanu, A.I., Han, Z., Paster, K., Pitis, S., Chan, H., Ba, J., 2022. Large language models are human-level prompt engineers, in: arXiv Preprint. Presented at the International Conference on Learning Representations (ICLR) 2023, p. 43. https://doi.org/10.48550/ARXIV.2211.01910

**Appendix S1: Impact of prompt variants on synthetic image quality and usability**

**Preliminary Prompts Tested**

The present study set out to generate realistic and ecologically plausible synthetic images of muskoxen using the diffusion-based model DALL-E 2. To this end, several text prompts were iteratively tested. Our goal was to simulate the aerial image acquisition conditions used in wildlife monitoring campaigns, especially capturing nadir views of muskoxen in snow-covered environments. The key prompts that were tested are summarized and the rationale for the final selected prompt is provided (see Table 1).

**Table 5.** Prompt variants tested and observed effects on image quality and ecological plausibility.

| Prompt | Observations [1] |
|---|---|
| *"A herd of muskoxen in the tundra seen from above"* | Animals appeared too small or distant, making individual identification difficult. Snowy backgrounds were rarely generated, and perspectives were inconsistent. |

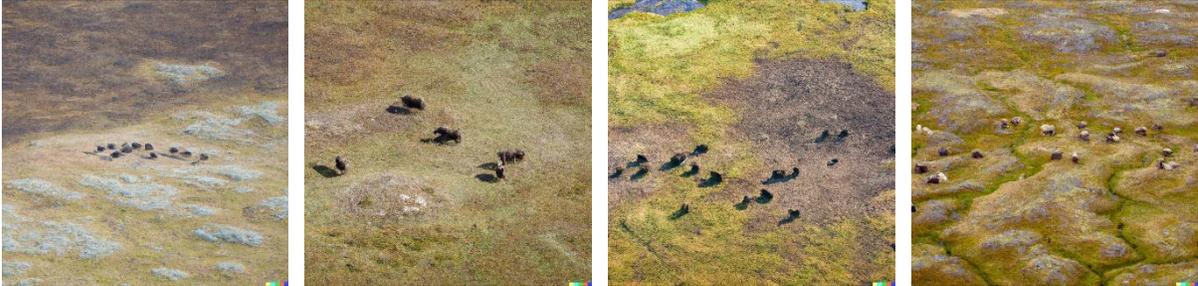

| Prompt | Observations |
|---|---|
| *"Drone photography of a herd of muskoxen in winter"* | Snowy scenes were often generated, but herds were unrealistically large, and individual animals were frequently distorted or shown in unnatural poses. |

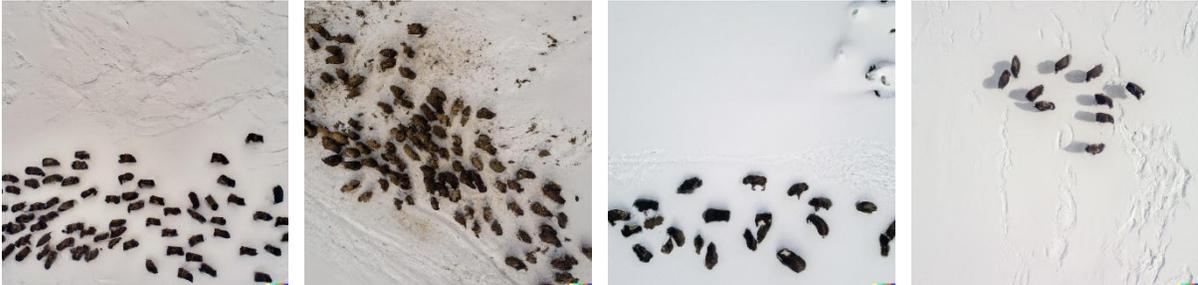

| Prompt | Observations |
|---|---|
| *"Realistic drone photography of a herd of muskoxen in winter"* | Added the word "realistic", which improved lighting and textures. However, images frequently lacked a nadir view and incorporated stylized or artistic renderings of animals. |

| Prompt | Observations [1] |
|---|---|

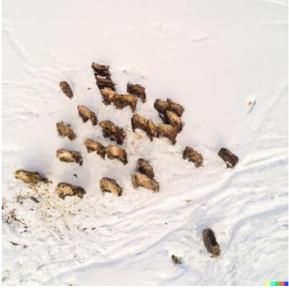 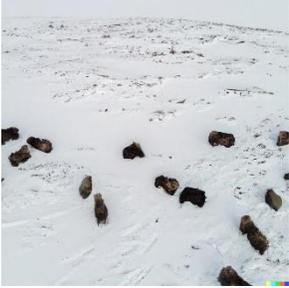 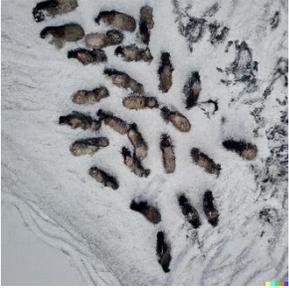 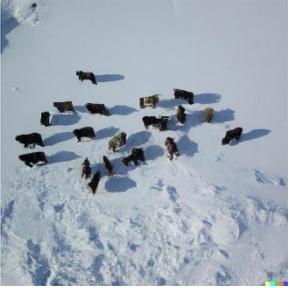

| "*A realistic aerial photography of a herd of muskoxen in winter seen from above*" | Image composition was enhanced; however, muskoxen frequently blended into the snowy or rocky background, making annotation unreliable or impossible. |
|---|---|

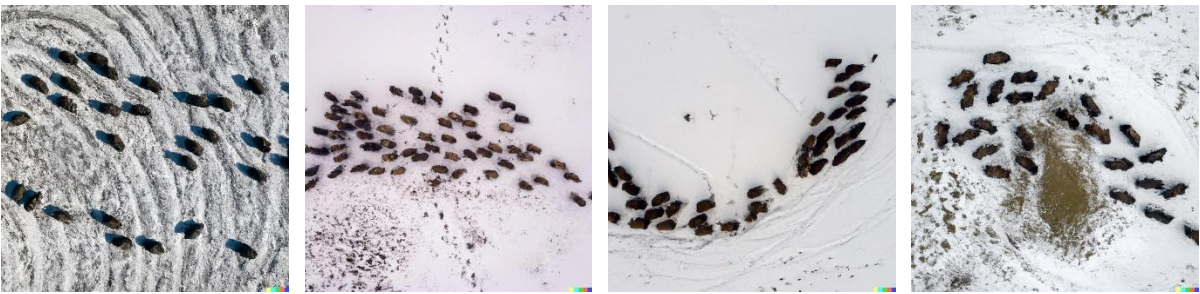

[1] Examples have been significantly downsized to fit the table. Zooming in is necessary to better appreciate the artifacts and quality issues.

**Retained Prompt**

Following an iterative testing process, the prompt "*Herd of muskoxen seen from above with a winter background, aerial imagery*" was selected as the most effective for generating synthetic images that are aligned with our case study. This prompt produced visually coherent outputs that closely matched the characteristics of real aerial survey imagery. Several aspects may have contributed to its effectiveness:

- **Viewpoint:** The phrase "*seen from above*" reliably produced a nadir-like perspective, which is crucial for replicating aerial survey conditions.
- **Scene realism:** Including "*winter background*" prompted the generation of snow-covered terrain, matching the environmental context of our study.
- **Stylistic alignment:** The term "*aerial imagery*" favored less stylized and more photorealistic renderings, thus avoiding the artistic distortions common in some diffusion outputs.
- **Appropriate group size:** By omitting specific herd counts, the prompt avoided unrealistic animal densities within a single image. In the few cases where excessive group sizes occurred, such images were easily identified and discarded during manual filtering.

- **Clarity of subjects:** Most importantly, this prompt yielded images in which muskoxen were distinct and appropriately sized for reliable manual annotation.

**Prompt Recommendations**

The following recommendations which are grounded in our own testing and iterative prompt evaluation, aim to guide the generation of high-quality synthetic data for ecological research and related applications.

1) **Begin with general text prompts and refine them iteratively**. Starting with simple, high-level descriptions allows for rapid prototyping and assessment of the model's output. As specific needs emerge, such as certain animal behaviors, environments, or camera angles, these prompts can be gradually adjusted to improve visual fidelity and ecological plausibility.

2) **Avoid overly specific or niche combinations of terms**. While it may be tempting to describe a highly detailed scene, extremely narrow prompts can produce unrealistic or inconsistent results if the model has not encountered similar examples during training. Balancing specificity with generality helps maintain output quality while capturing essential features of the target scene.

3) **Human review remains essential**. Despite ongoing improvements in generative models, automated tools for evaluating image quality or content accuracy are still limited in scope and reliability. For scientific applications, particularly in domains where fine-grained ecological or taxonomic detail matters, visual validation by domain experts is critical.

4) **Plan for variability in output quality**. Even well-crafted prompts can yield inconsistent or flawed results, especially when generating large image sets. Manual filtering remains an important step, though it can be supplemented with automated quality control tools where available. Until generative models become more robust and context-aware, incorporating quality assurance measures into the data generation pipeline will be necessary.

**Appendix S2: Model hyperparameters used for ODM training**

To optimize object detection model (ODM) performance, multiple training iterations were conducted, with parameters adjusted in each iteration to refine muskox detection and counting. Given that the ODMs were designed to detect a single object class (muskoxen), the F1 score was preferred over mean average precision as the primary evaluation metric. To minimize artifacts and distortions, bilinear interpolation resampling was used for image resizing. Images used for ODM training and validation were resized to ensure that muskoxen had an average length of approximately 100 pixels, while those used for testing were resized to an average length of 70 pixels. During patch creation, partial labels were retained if at least 50% of their surface area overlapped with the patch. Additionally, global CNN training, where all layers are adjusted simultaneously, outperformed sequential training and was therefore adopted for ODM training. Several preprocessing parameters, including patch width (512 pixels), height (512 pixels), overlap (256 pixels), number of epochs (300), learning rate (0.001), and HerdNet's metric radius (30 pixels), remained consistent throughout the optimization process. The classic augmentations from the Python library *Albumentations* used for ODM training (Table 2) were also kept consistent across all trained ODMs.

**Table 6.** Classic augmentations implemented using the Python library *Albumentations*.

| Operation order | Augmentation | Parameters |
|---|---|---|
| 1 | RandomBrightnessContrast | brightness_limit = (-0.45, 0.55) |
| | | contrast_limit = (-0.60, 1.00) |
| | | p = 0.5 |
| 2 | HueSaturationValue | hue_shift_limit = (-5, 5) |
| | | sat_shift_limit = (-50, 50) |
| | | val_shift_limit = (-20, 20) |
| | | p = 0.5 |
| 3 | Flip | p = 0.5 |
| 4 | RandomRotate90 | p = 0.5 |
| 5 | RandomScale | scale_limit = (-0.15, 0.15) |
| | | interpolation = cv2.INTER_LINEAR |
| | | p = 0.5 |
| 6 | Blur | blur_limit = (2, 4) |
| | | p = 0.5 |
| 7 | PadIfNeeded | min_height = 512 |
| | | min_width = 512 |
| | | border_mode = 0 |
| | | always_apply = True |
| 8 | CenterCrop | Height = 512 |
| | | Width = 512 |
| | | always_apply = True |
| 9 | Normalize | always_apply = True |

**Appendix S3: Metric values obtained for each object the detection model created**

**Table 7.** Performance results obtained for all metrics (AP: average precision, MAE: mean absolute error, MSE: mean squared error, RMSE: root mean squared error, precision, recall, F1 score) for all object detection models created (BL: baseline model, FS1-5: few-shot models, ZS1-5: zero-shot models).

| Model | Fold | AP | MAE | MSE | RMSE | Precision | Recall | F1 score |
|-------|------|------|------|-------|------|-----------|--------|----------|
| BL | 1 | 0.89 | 0.54 | 1.24 | 1.12 | 0.94 | 0.90 | 0.92 |
| | 2 | 0.87 | 0.56 | 1.28 | 1.13 | 0.93 | 0.90 | 0.92 |
| | 3 | 0.81 | 0.76 | 2.27 | 1.51 | 0.92 | 0.84 | 0.88 |
| | 4 | 0.82 | 0.72 | 2.15 | 1.47 | 0.94 | 0.85 | 0.89 |
| | 5 | 0.75 | 0.95 | 3.93 | 1.98 | 0.94 | 0.78 | 0.86 |
| FS1 | 1 | 0.78 | 0.91 | 3.76 | 1.94 | 0.96 | 0.80 | 0.87 |
| | 2 | 0.87 | 0.65 | 1.85 | 1.36 | 0.90 | 0.91 | 0.91 |
| | 3 | 0.83 | 0.69 | 2.21 | 1.49 | 0.94 | 0.86 | 0.90 |
| | 4 | 0.86 | 0.63 | 1.72 | 1.31 | 0.93 | 0.89 | 0.91 |
| | 5 | 0.87 | 0.60 | 1.60 | 1.27 | 0.92 | 0.90 | 0.91 |
| FS2 | 1 | 0.92 | 0.51 | 1.20 | 1.09 | 0.92 | 0.94 | 0.93 |
| | 2 | 0.86 | 0.68 | 2.23 | 1.49 | 0.88 | 0.93 | 0.90 |
| | 3 | 0.86 | 0.66 | 2.16 | 1.47 | 0.90 | 0.90 | 0.90 |
| | 4 | 0.88 | 0.59 | 1.45 | 1.21 | 0.92 | 0.91 | 0.91 |
| | 5 | 0.80 | 0.84 | 3.14 | 1.77 | 0.96 | 0.81 | 0.88 |
| FS3 | 1 | 0.86 | 0.64 | 1.65 | 1.29 | 0.89 | 0.91 | 0.90 |
| | 2 | 0.86 | 0.66 | 2.16 | 1.47 | 0.88 | 0.93 | 0.91 |
| | 3 | 0.86 | 0.63 | 1.71 | 1.31 | 0.91 | 0.90 | 0.90 |
| | 4 | 0.91 | 0.54 | 1.29 | 1.14 | 0.91 | 0.94 | 0.93 |
| | 5 | 0.92 | 0.57 | 1.73 | 1.32 | 0.90 | 0.95 | 0.92 |
| FS4 | 1 | 0.91 | 0.50 | 1.09 | 1.04 | 0.92 | 0.93 | 0.93 |
| | 2 | 0.87 | 0.61 | 1.59 | 1.26 | 0.90 | 0.92 | 0.91 |
| | 3 | 0.85 | 0.66 | 1.91 | 1.38 | 0.93 | 0.87 | 0.90 |
| | 4 | 0.87 | 0.56 | 1.30 | 1.14 | 0.92 | 0.92 | 0.92 |
| | 5 | 0.93 | 0.59 | 1.70 | 1.30 | 0.89 | 0.95 | 0.92 |
| FS5 | 1 | 0.86 | 0.63 | 1.68 | 1.30 | 0.91 | 0.90 | 0.90 |
| | 2 | 0.87 | 0.57 | 1.36 | 1.17 | 0.92 | 0.91 | 0.91 |
| | 3 | 0.88 | 0.68 | 2.02 | 1.42 | 0.88 | 0.92 | 0.90 |
| | 4 | 0.87 | 0.60 | 1.61 | 1.27 | 0.90 | 0.93 | 0.91 |
| | 5 | 0.87 | 0.59 | 1.54 | 1.24 | 0.90 | 0.92 | 0.91 |
| ZS1 | 1 | 0.71 | 1.85 | 20.94 | 4.58 | 0.72 | 0.78 | 0.75 |
| | 2 | 0.74 | 1.55 | 9.84 | 3.14 | 0.79 | 0.77 | 0.78 |
| | 3 | 0.69 | 1.82 | 11.61 | 3.41 | 0.74 | 0.74 | 0.74 |
| | 4 | 0.72 | 1.62 | 11.63 | 3.41 | 0.78 | 0.76 | 0.77 |
| | 5 | 0.74 | 1.71 | 11.44 | 3.38 | 0.75 | 0.77 | 0.76 |
| ZS2 | 1 | 0.75 | 1.33 | 6.65 | 2.58 | 0.82 | 0.79 | 0.80 |
| | 2 | 0.78 | 1.46 | 7.33 | 2.71 | 0.74 | 0.84 | 0.79 |
| | 3 | 0.72 | 1.31 | 5.73 | 2.40 | 0.85 | 0.75 | 0.80 |
| | 4 | 0.76 | 1.46 | 7.22 | 2.69 | 0.77 | 0.81 | 0.79 |
| | 5 | 0.73 | 1.26 | 5.41 | 2.33 | 0.85 | 0.77 | 0.81 |
| ZS3 | 1 | 0.79 | 1.23 | 5.23 | 2.29 | 0.82 | 0.82 | 0.82 |
| | 2 | 0.82 | 1.01 | 3.83 | 1.96 | 0.86 | 0.85 | 0.85 |
| | 3 | 0.74 | 1.16 | 4.73 | 2.18 | 0.91 | 0.76 | 0.83 |
| | 4 | 0.75 | 1.25 | 5.81 | 2.41 | 0.89 | 0.76 | 0.82 |
| | 5 | 0.78 | 0.96 | 3.12 | 1.77 | 0.90 | 0.81 | 0.85 |
| ZS4 | 1 | 0.78 | 0.89 | 2.82 | 1.68 | 0.90 | 0.82 | 0.86 |

| Model | Fold | AP | MAE | MSE | RMSE | Precision | Recall | F1 score |
|-------|------|------|------|------|------|-----------|--------|----------|
|       | 2    | 0.79 | 1.01 | 3.67 | 1.92 | 0.90      | 0.80   | 0.85     |
|       | 3    | 0.78 | 1.02 | 3.83 | 1.96 | 0.90      | 0.80   | 0.85     |
|       | 4    | 0.77 | 1.00 | 3.68 | 1.92 | 0.87      | 0.81   | 0.84     |
|       | 5    | 0.76 | 1.16 | 4.73 | 2.18 | 0.88      | 0.78   | 0.83     |
| ZS5   | 1    | 0.74 | 1.05 | 4.23 | 2.06 | 0.90      | 0.77   | 0.83     |
|       | 2    | 0.80 | 1.02 | 4.04 | 2.01 | 0.88      | 0.82   | 0.85     |
|       | 3    | 0.84 | 0.97 | 3.59 | 1.89 | 0.83      | 0.88   | 0.85     |
|       | 4    | 0.70 | 1.26 | 6.00 | 2.45 | 0.94      | 0.72   | 0.81     |
|       | 5    | 0.70 | 1.20 | 5.09 | 2.26 | 0.90      | 0.73   | 0.81     |

## Appendix S4: Complementary analysis including recent generative models

### Image generation

The same prompt that was used to generate the DALL-E 2 images was used ("Herd of muskoxen seen from above with a winter background, aerial imagery") to generate synthetic images using these two recent models. These images were then sorted to retain only realistic images using the same criteria as those used for the selection of DALL-E 2 synthetic images (nadir, snowy background). The image selection results are presented in Table 1.

**Table 4.** Number of synthetic images generated by three models and fractions of realistic image selected for analysis.

| Model | Number of generated synthetic images | Number of selected synthetic images | Fraction of selected synthetic images |
|---|---|---|---|
| DALL-E 2 | 1000 | 160 | 16% |
| DALL-E 3 | 612 | 160 | 26% |
| Gemini 2.5 Flash Image | 200 | 1 | 0.5% |

Results show that Gemini was unable to generate enough realistic images to reach the number required in the analysis. Given the very low rate of realistic images selected, we stopped image production at 200. On the other hand, DALL-E 3 showed a higher percentage of selected images than DALL-E 2, demonstrating an improvement in the realism of the images generated for a survey context. Figure 1 illustrates some examples of synthetic images generated by the three models that show the realism and artifacts produced by each model.

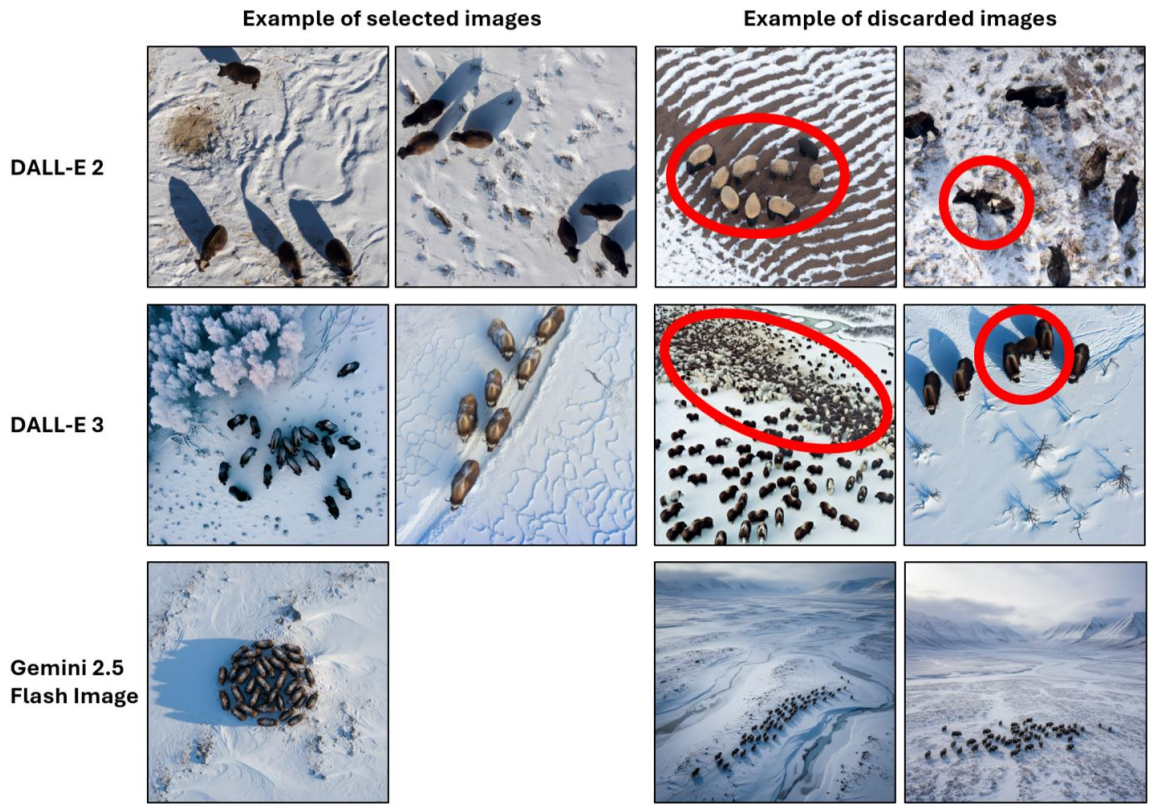

**Figure 1.** Examples of selected and discarded synthetic images generated by three models: DALL-E 2, DALL-E 3, and Gemini 2.5 Flash Image. Discarded images exhibit colour (DALL-E 2), physiology (DALL-E 2, DALL-E 3), background (DALL-E 3), and viewing angle (Gemini 2.5 Flash Image) anomalies.

### Model performance

Images selected from DALL-E 3 models were analyzed using the same approach as the DALL-E 2 images, using the same test dataset. Figure 2 illustrates the accuracy, recall, and F1-score of the model trained on all synthetic images generated by DALL-E 3 compared to the analyses performed using DALL-E 2 images.

Results show a similar F1 score between models trained using the same number of images generated by DALL-E 2 (ZS5) and DALL-E 3 (ZS5*). It demonstrates a similar model's ability to balance precision and recall. However, ZS5* shows lower precision than ZS5, while recall is higher for ZS5* compared to ZS5.

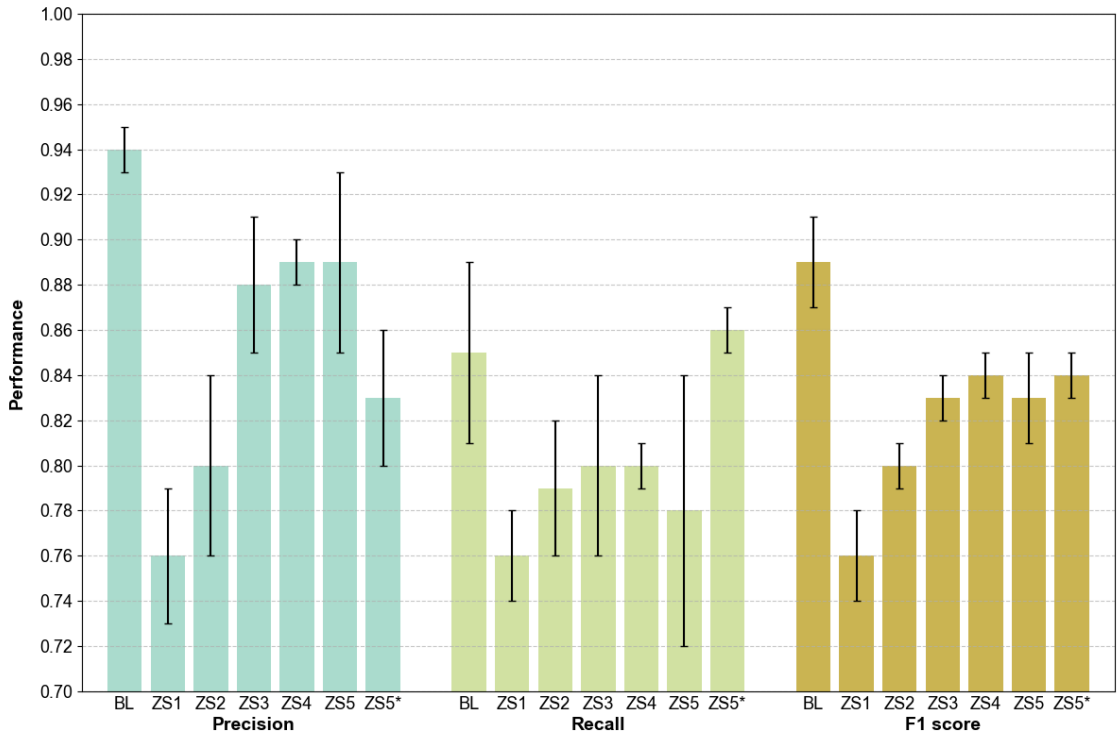

**Figure 2.** Comparison of performance metrics for the baseline (BL) and the zero-shot (ZS1-ZS5) models, including precision, recall, and F1 score. Standard deviation obtained from 5-fold cross-validation is visually represented as a vertical interval at the top of each bar. ZS1-ZS5 include synthetic images generated from DALL-E 2 whereas ZS5* includes images generated from DALLE-3.

Overall, these complementary analyses indicate that recent models, such as Gemini (Gemini 2.5 Flash Image), still lack the ability to produce realistic images of a wildlife survey context. On the other hand, other models such as DALL-E 3 show an improvement in the realism of the images produced in this context (26% of images retained) compared to DALL-E 2 (16% of images retained). However, these improvements do not translate into clear improvement in model performance.